\documentclass[letterpaper,11pt]{amsart}
\pdfoutput=1
\usepackage{latexsym}
\usepackage{amssymb}
\usepackage{amsfonts}
\usepackage{amsmath}
\usepackage{amsthm}
\usepackage[pdftex]{graphicx}

\newcommand{\fat}[1]{\mathbb{#1}}
\newcommand{\hsm}[1]{\mathcal{#1}}
\newcommand{\ZZ}{\fat{Z}}
\newcommand{\OO}{\hsm{O}}

\newcommand{\IFF}{\,\Leftrightarrow\,}
\newcommand{\minus}{\smallsetminus}

\newcommand{\inv}{^{{\scriptscriptstyle -1}}}

\newcommand{\PR}[2]{\mathrm{Pr}_{\scriptscriptstyle{#1}}\left(#2\right)} 

\newcommand{\same}[1]{\stackrel{\scriptscriptstyle{#1}}{\sim}}
\newcommand{\card}[1]{\left|#1\right|}

\newcommand{\med}{\mathrm{med}}
\newcommand{\Poc}[1]{\hsm{P}\left(#1\right)}
\newcommand{\PocA}{\hsm{P}\left(\fat{A}\right)}
\newcommand{\supp}[1]{\mathrm{supp}\left(#1\right)}

\renewcommand{\square}{\Delta}
\newcommand{\ch}[1]{\mathrm{ch}\!\left(#1\right)}

\theoremstyle{definition}\newtheorem{thm}{Theorem}[section]

\newtheorem{prop}[thm]{Proposition}
\newtheorem{lemma}[thm]{Lemma}
\newtheorem{defn}[thm]{Definition}
\newtheorem{example}[thm]{Example}

\theoremstyle{remark}\newtheorem{remark}[thm]{Remark}

\begin{document}
\title{A Formal Approach to Modeling the Memory of a Living Organism}
\author[D.\ P.~Guralnik]{Dan P.\ Guralnik}
\address{Dept.\ of Mathematics\\ University of Oklahoma\\ Norman, OK 73019}
\email{dan.guralnik@ou.edu}
\begin{abstract} We consider a living organism as an observer of the evolution of its environment recording sensory information about the state space $X$ of the environment in real time. Sensory information is sampled and then processed on two levels. On the biological level, the organism serves as an {\it evaluation mechanism} of the subjective relevance of the incoming data to the observer: the observer assigns {\it excitation values} to events in $X$ it could recognize using its sensory equipment. On the algorithmic level, sensory input is used for updating a database -- the memory of the observer -- whose purpose is to serve as a geometric/combinatorial model of $X$, whose nodes are weighted by the excitation values produced by the evaluation mechanism. These values serve as a guidance system for deciding how the database should transform as observation data mounts.

We define a searching problem for the proposed model and discuss the model's flexibility and its computational efficiency, as well as the possibility of implementing it as a dynamic network of neuron-like units. We show how various easily observable properties of the human memory and thought process can be explained within the framework of this model. These include: reasoning (with efficiency bounds), errors, temporary and permanent loss of information. We are also able to define general learning problems in terms of the new model, such as the language acquisition problem.
\end{abstract}


\maketitle

{\it\hfill Dedicated to the memory of my father,}

{\it\hfill  Peter J. Guralnik, and all his mice and rats.}

\bigskip

\section{Introduction}\label{section:introduction}
\subsection{General considerations.}\label{subsection:general considerations} The structure of memory in living organisms is generally perceived as extremely complex and extremely efficient at the same time. Complex, because the sheer number of physiological structure elements comprising the nervous system of a {\it rattus rattus}, say -- not to mention {\it homo sapiens} -- seems to exclude the possibility of direct piece-by-piece analysis. Efficient, because of the capability to respond to unpredicted input signals in real time, frequently with desirable outcome.

Different species seem to exhibit different capacities for this kind of thinking, and the reason for this remains unclear. By this we mean that, although biologists have a lot to say about the evolution of the central nervous system, and in spite of our ability to roughly map the brain and say which parts of it are in charge of which functions, one major problem remains completely open: {\it what is the algorithmic structure underlying the phenomenon we call `intelligence'?}

Answering this question adequately is one of the dream goals of the field of Artificial Intelligence, whose ultimate goal -- one should na\"ively guess -- is to construct machines capable of humanoid reasoning, however much faster and more accurate.

Developing mathematical tools for describing the principles governing our thought process and the development of our minds may turn out useful for improving our understanding of learning and teaching, and may seriously impact psychology. Describing the relationship between language formation and acquisition, structure of memory, logical thinking and psychology may finally be within our grasp.\\

Considering the wide variety of applications it becomes desirable to formulate the answer to our main question in a way that is independent of its physical realization (e.g. the biophysics of the brain). We will refer to this as the {\it `invariance principle'}.

The invariance principle defines an objective measure of adequacy for our attempted answers. Indeed, if invariance is observed, then whenever an abstract model of the mind is offered, it becomes possible to study multiple realizations, which gives us the chance to search for the most efficient ones; if no efficient realization exists, we discard the model. On the other hand, if a model is tied to a specific physical realization, then its predictive power is automatically limited by our understanding of that realization. The latter usually being incomplete also puts us in jeopardy of using an intrinsically inconsistent model for predictions.\\

In this work we introduce an approach based on considering memory as an algorithmic structure: a database, where data is stored, together with a set of algorithms in charge of maintaining the structure and retrieving stored information. By `memory' we mean the broadest possible interpretation of the term: all information observed and retained by the living organism in the course of its lifetime, together with the procedures handling all this data.

Though contributing to the feeling that analysis of such a `memory' is hopelessly hard, this interpretation is a necessity dictated by the invariance principle. In the case of animals (including humans) it was Evolution that shaped our memory management system and distributed its functions among the many scales and subsystems of our organisms, solving this problem on the way, in the course of billions of years. It is then only reasonable to assume that when it comes to producing a {\it homunculus} of our own making, we will need to face the same challenges. Therefore there is no point in restricting our theoretical framework.

It is unclear how memory structures of living organisms are maintained and what principles are responsible for the seeming high efficiency of their recording and retrieval algorithms. In the preceding sentence, the word `seeming' is used because, in contrast to the general feeling of awed amazement at the capabilities of human and animal brains, frequent everyday observations of humans strongly suggest that said brains possess some rather discouraging inherent flaws. For example, trouble recovering useful information while recalling seemingly irrelevant data with ease ({\it ``what was that formula for $\cos 2\theta$?''} vs. {\it ``can not get that idiotic tune out of my head!''}), failure to recognize phenomena that are supposedly well understood ({\it ``why didn't *I* think about that?!''}), as well as difficulties in processing logically complex statements (e.g. {\it he thinks that she thinks that he knows that she heard that he thinks she believes that he loves someone else}), -- are all everyday common examples of what we normally perceive as `glitches' of our memory system. We feel that if only we could merge ourselves with a powerful computer, all these inaccuracies will be gone.

One possible reason for this feeling is that we tend to see our memory management structure as the result of a very successful evolutionary process. This idea makes it too easy for us to focus only on the desirable manifestations of memory structuring when trying to replicate nature's achievements, while all the undesirable effects we observe on a daily basis are written off as malfunctions of a complicated analog system (our organism), occurring due to fatigue and/or some other physiological impairment.

On the other hand, one could also argue that the high frequency of malfunctions is the manifestation of a set of principles governing the way in which our memory resources are managed. Indeed, since {\it any} organizing principle in fact constitutes a restriction on the set of admissible structures (that is, some database structures will never be realized because they violate said principle), it is reasonable to expect many admissible structures to react inadequately to some unpredictable situations presented by the environment. Therefore, the optimist will regard all the failings of our memory storage system as hints to how such a system is structured, and the quality of a model should be judged not only by its computational efficiency, but more by its ability to explain the functional role of errors in the system, how errors occur, which errors occur more frequently than others, etc.

\subsection{Stating the problem.}\label{subsection:stating the problem}
It is time to state our goals in a more committing and formal fashion. We consider the organism's memory as a record of its observations of the evolution of the environment (participation by the organism is not ruled out). The environment as a whole has an associated space of states -- denoted henceforth by $X$, -- and every organism $\hsm{O}$ maintains a database $\Gamma$ whose structure and content correspond to that observer's perception of properties of $X$, as determined by the available sensory equipment. Thus, we consider the organism $\hsm{O}$ as a mediator between the environment and the database $\Gamma$.

Recall that a database normally has the structure of a graph (or {\it network}), with nodes carrying additional information (or {\it content}). Updating the database may involve altering the content of a node or nodes, as well as adding new nodes or erasing redundant ones.

When an organism $\hsm{O}$ with memory structure $\Gamma$ makes an observation about $X$ (receives new input from the environment), this observation will be evaluated (this means $\Gamma$ is being read), possibly resulting in an updating procedure and replacing $\Gamma$ by a new structure $\Gamma'$. Mathematically speaking, this means our model should have an underlying structure of a category $\hsm{C}$, whose objects belong to the class of admissible databases, and the morphisms (also known as {\it arrows}) describe the various possibilities of transforming one structure into the other. The structure (e.g. algebraic, topological, other...) of this category corresponds to the idea of a set of `governing principles' we have just discussed.

Recall that a category $\hsm{C}$ consists of a class of objects $\mathrm{ob}\hsm{C}$, a set $\hsm{C}(A,B)$ of morphisms $f:A\to B$, defined for every ordered pair of objects $A,B\in\mathrm{ob}\hsm{C}$, and an operation (called composition), defined for each ordered triple $A,B,C$ of objects --
\begin{displaymath}
	\left\{\begin{array}{rccl}
		\circ:&\hsm{C}(A,B)\times\hsm{C}(B,C)&\longrightarrow&\hsm{C}(A,C)\\
		&f\times g&\mapsto&g\circ f
	\end{array}\right.
\end{displaymath}
Composable morphisms are said to be {\it compatible}. All the above must satisfy the following requirements:
\begin{itemize}
	\item For every  $A\in\mathrm{ob}\hsm{C}$, the set $\hsm{C}(A,A)$ contains a distinguished element denoted $\mathrm{id}_A$;
	\item For every  $A,B\in\mathrm{ob}\hsm{C}$ and any $f\in\hsm{C}(A,B)$ one has $f\circ\mathrm{id}_A=\mathrm{id}_B\circ f=f$;
	\item For every triple of compatible morphisms $f,g,h$, one has $h\circ(g\circ f)=(h\circ g)\circ f$.
\end{itemize}
Contemplating the meaning of the axioms of a category in our context, one may regard a morphism between two objects (databases) as a means for comparing them. One may want to measure the improvement resulting from updating $\Gamma$ into $\Gamma'$ (see above), or one could be interested in measuring the differences in how two distinct observers perceive a common environment. Finally, one may want to measure the discrepancy between an observer's perception and the objective reality presented by the environment. Whether or not these are possible in any useful sense depends on the structure of the category underlying our modeling method.\\

The motivation for our construction has two sources: one is Shannon's idea of {\it entropy} introduced in \cite{[Shannon]}, and the other is the idea of {\it spaces with walls} introduced by Haglund and Paulin in \cite{[Walls]}. The marriage of the two produces the information-theoretic approach to constructing databases based on binary observations, which we shall describe right now.

It is natural to assume the state space $X$ of the observed environment is a topological space endowed with the corresponding Borel $\sigma-$algebra $\hsm{B}$ and a probability measure $\mu$, making it possible to consider the probability of an event in $X$. Let us fix a moment in time and an observer $\hsm{O}$ and assume that the sensory equipment available to $\hsm{O}$ produces binary output. Assume there are only finitely many sensors available to $\hsm{O}$. Imagine an angel (as opposed to the notion of a daemon, frequently used in the literature to explain various notions of entropy) who is in charge of recording the output from these sensors and keeping it in order.

The angel is absolutely objective: it does not have a preference to any kind of data generated by the sensors, and its sole and sacred responsibility is to record the data as accurately as possible. The angel is all-knowing: for each sensor, it knows precisely which inputs (states of the environment) produce which output for that sensor.

Thus, in the angel's notebook, each sensor will correspond to a pair of complementary subsets of $X$, while the totality of all information that $\OO$ is capable of producing then becomes a finite family $H$ of subsets of $X$, which is closed under complementation. Even angels have limited powers, so $H\subseteq\hsm{B}$.

Associated with $H$ is a partition of $X$: given $x,y\in X$, write $x\same{H}y$ if and only if $x\in h$ implies $y\in h$ for every $h\in H$. The relation $(\same{H})$ is then an equivalence relation whose induced partition (denote it by $\hsm{P}(H)$) is the join (the coarsest common refinement) of the binary partitions corresponding to the individual sensors. It is important that the angel keep this partition on the record: the observer $\hsm{O}$ is incapable of discerning between two events belonging to the same element of this partition. For this reason, the elements of $\hsm{P}(H)$ will be called {\it $H$-visible states}.

From the point of view of the angel, it is now possible to determine the amount of information that $\hsm{O}$ has about $X$: this task is equivalent to computing Shannon's entropy of $\hsm{P}(H)$. Unfortunately, this result is not meaningful for the observer $\hsm{O}$: Shannon's entropy is approximated by the minimum expectation of the number of {\it arbitrary} binary questions one needs to ask in order to determine the position of a point of $X$ with respect to $\hsm{P}(H)$, but $\hsm{O}$ only has the questions from $H$ at his disposal, so the minimum computed by $\hsm{O}$ may end up much higher than that computed by the all-knowing angel, who is surely capable of asking {\it any} question from the list $\hsm{B}$.

Despite the differences in computational ability, both the observer and the angel are interested in having the probabilities $\mu(P)$ of each element $P\in\hsm{P}(H)$ recorded somewhere. Thus, we have content for our database, but no graph to put it in. This is easily repaired. Define a graph $\Gamma_H$ to have $\hsm{P}(H)$ for its set of vertices, where two vertices $P,Q\in\hsm{P}(H)$ are joined by an edge if and only if there exists precisely one $h\in H$ satisfying $P\subseteq h$ and $Q\subseteq h^c$ (here and on, $h^c=X\minus h$). This construction is just a special case of the construction of the graph dual to a space with walls. Note that $\Gamma_H$ is necessarily connected and bipartite. Perhaps it is time for an example:

\begin{figure}[t]
	\centering{\includegraphics[width=\textwidth]{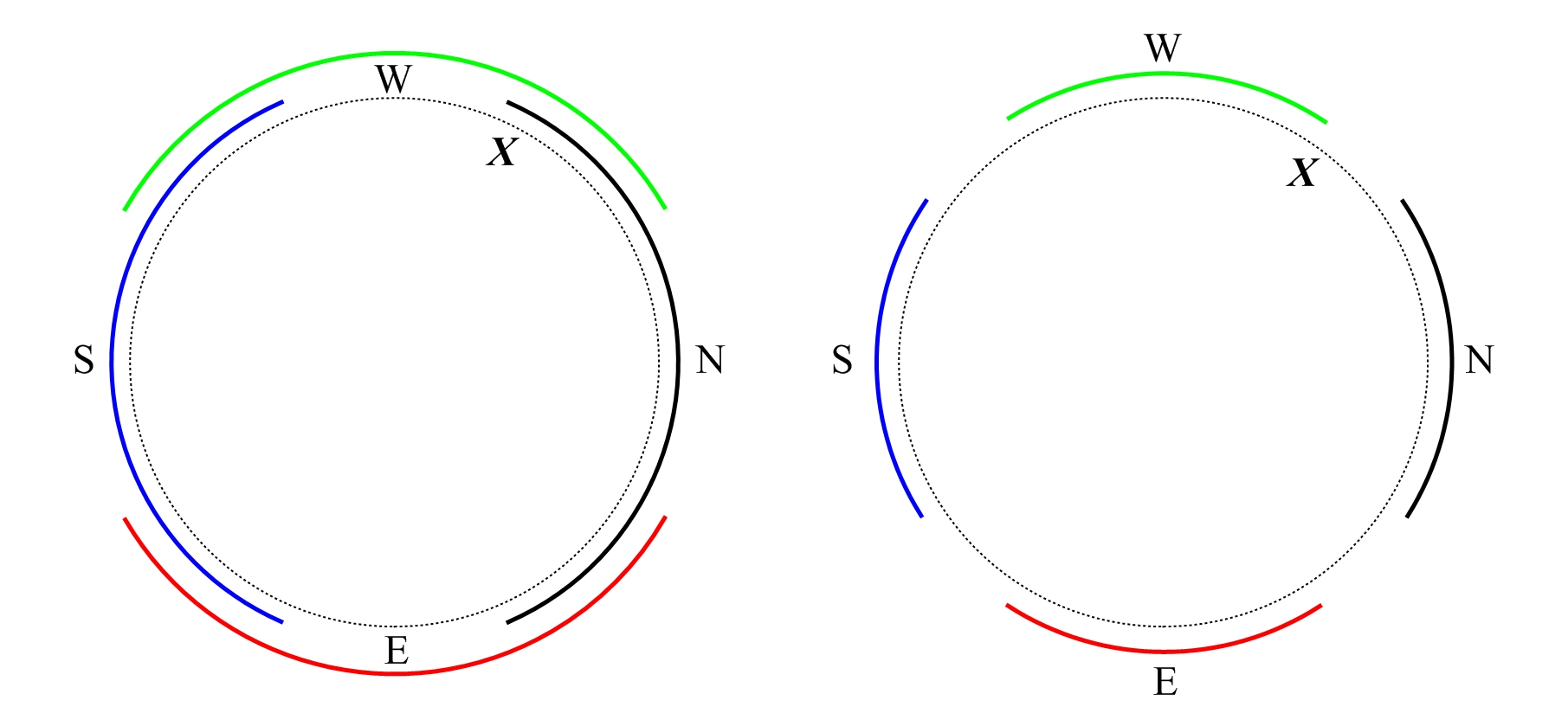}}
	\caption{\protect\scriptsize Observing a compass (1). Two versions of the system $H$ of example \ref{example:compass0}.\protect\normalsize
		\label{figure:compass0}}	
\end{figure}

\begin{figure}[t]
	\centering{\includegraphics[width=\textwidth]{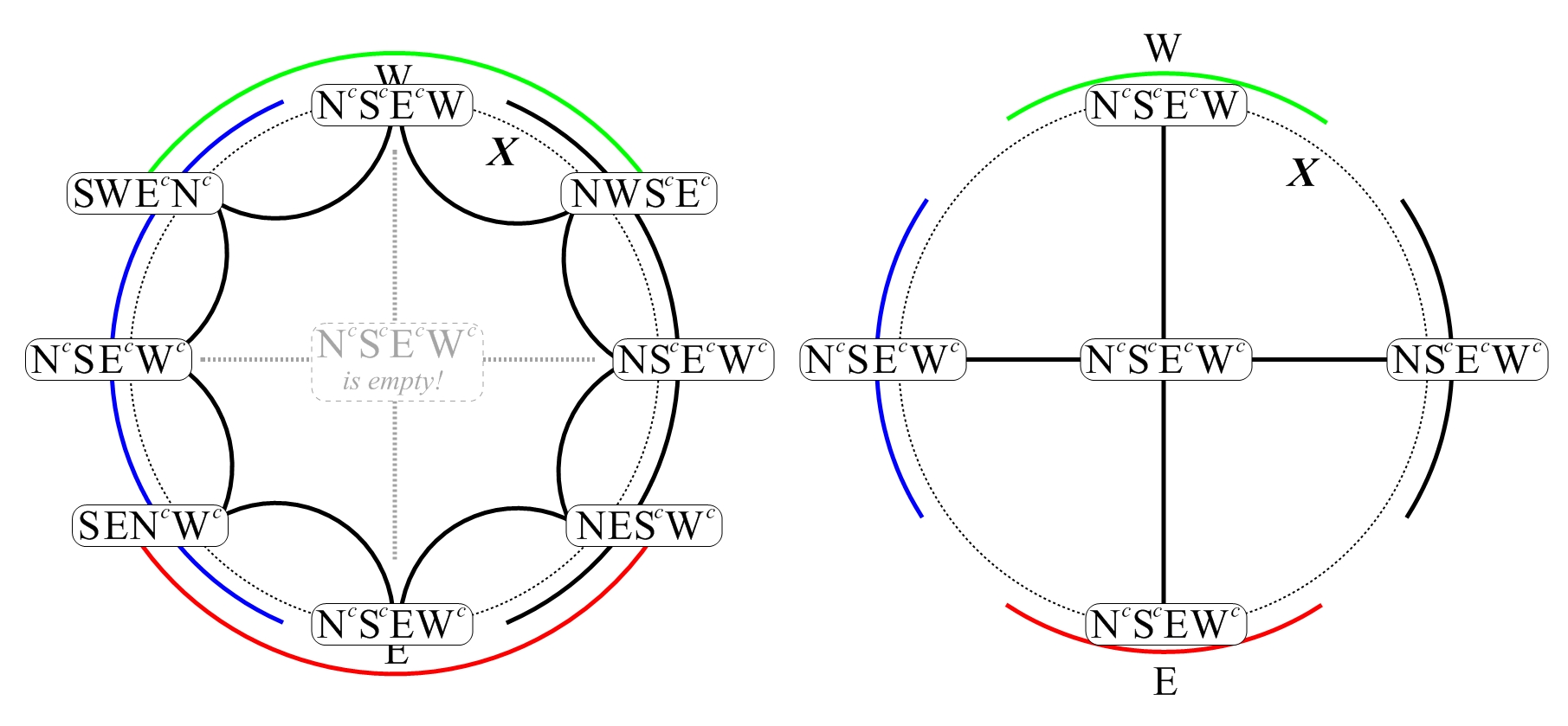}}
	\caption{\protect\scriptsize Observing a compass (1). The graphs $\Gamma_H$ for the two types of the system $H$ from example \ref{example:compass0} -- compare with fig. \ref{figure:compass0}.\protect\normalsize
		\label{figure:compass05}}	
\end{figure}

\begin{example}[observing a compass, part 1]\label{example:compass0} Consider a person $\hsm{O}$ observing a compass. The space of states of the needle of the compass can be modeled by the unit circle $X=\fat{S}^1$, thought of as a subset of the complex numbers $\fat{C}$, with the number $\mathbf{1}$ corresponding to precise North, and $\mathbf{i}$ corresponding to West.

Now, imagine our observer being able to ask the questions ``Is the needle pointing North/South/West/East?'' (and their complements, of course), and let us model the positive answer sets -- denoted by $N,S,W$ and $E$, respectively -- by open intervals:
\begin{eqnarray*}
	N&=&\left\{e^{i\theta}\,\Big|\,
		\theta\in(-\epsilon,\epsilon)\right\}\\
	S&=&\left\{e^{i\theta}\,\Big|\,
		\theta\in(\pi-\epsilon,\pi+\epsilon)\right\}\\
	W&=&\left\{e^{i\theta}\,\Big|\,
		\theta\in(\frac{\pi}{2}-\epsilon,\frac{\pi}{2}+\epsilon)\right\}\\
	E&=&\left\{e^{i\theta}\,\Big|\,
		\theta\in(-\frac{\pi}{2}-\epsilon,-\frac{\pi}{2}+\epsilon)\right\}\,,
\end{eqnarray*}
where $\epsilon\in(0,\pi/2)$ is a number (in some sense characterizing the quality of the observations being made: a smaller $\epsilon$ means better precision). The observer may, initially be unaware or undecided regarding the value of $\epsilon$, so that two types of situations may occur: one with $\epsilon\leq\pi/4$ and the other with $\epsilon>\pi/4$ see figure \ref{figure:compass0}. We set \[H=\left\{N,S,W,E,N^c,S^c,W^c,E^c\right\}\]
and ask the reader to verify the pictures of the corresponding graphs $\Gamma_H$ presented in figure \ref{figure:compass05}. Observe how different the two graphs are. Does one of them resemble $X$ more than the other? Oddly enough, it is the {\it lower} quality observation that provided the better picture. How come?
\end{example}

The preceding example illustrates an idea central to this paper: memory is (or at least {\it should be}) a geometric/combinatorial model (here, a graph) of the state space $X$, marked with additional information (content at the vertices). It seems reasonable to also mark a point of this geometric model (in $\Gamma_H$ that would mean a vertex, or in other words, an $H$-visible state) to represent the observer's conjecture about the current state of $X$.

The categorical point of view discussed earlier will now work as follows: adding sensors to $\hsm{O}$ corresponds to an inclusion of the list $H$ in a longer list $\tilde H$, creating a refinement $\hsm{P}(\tilde H)$ of the partition $\hsm{P}(H)$ and hence a map of the graph $\Gamma_{\tilde H}$ onto the graph $\Gamma_H$ induced by sending every $\tilde H$-visible state to the unique $H$-visible state containing it. Thus the category $\hsm{C}$ for our modeling problem may be taken to have weighted connected bipartite graphs, and an arrow from an object $\Gamma$ to an object $\Gamma'$ is a surjective graph morphism (edges may be contracted) from $\Gamma'$ onto $\Gamma$.

A major problem standing in the way of developing this approach further is that $\Gamma_H$ can be quite arbitrary on the large scale, so that searching $\Gamma_H$ may turn out to be computationally unfeasible as the observer $\hsm{O}$ comes to possess more and more sensors and $H$ grows in size accordingly.

The situation is even worse than that: we have just assumed too much. First of all, the angel's job is {\it only} to do the book-keeping for $\hsm{O}$. In our interpretation of memory, this means the angel has no business representing the observer's questions by objectively defined complementary pairs of subsets of $X$: $\hsm{O}$ will most probably {\it never} have that much information about {\it any} of its sensors. This angel has to go home then, and we need to face the fact that the observer has to maintain $\Gamma_H$ and its content on his own, without any prior knowledge about the structure of $H$ and without any assurance that the content of the vertices of $\Gamma_H$ is objectively correct: $\hsm{O}$ is only able to {\it sample} states from $X$ by making repeated observations, so the probabilities recorded at the vertices may be very different from the objective ones.

To deliver the final blow, consider this: even if there is a way for $\hsm{O}$ to magically keep track of the objective structure of $\Gamma_H$, the observer may still be required to restructure $\Gamma_H$ numerous times as time evolves. For example, adding a vertex to $\Gamma_H$ where he initially thought there was none: suppose that, up to time $t=t_0$, $\hsm{O}$ has never sampled a state for which both questions $a$ and $b$ had a positive answer; as a result, up until time $t=t_0$, no vertex in $\Gamma_H$ listing a positive answer to $a$ lists a positive answer to $b$; if at time $t_0$ the observer suddenly makes the observation that $x\in a\cap b$ for some state $x\in X$, then $\Gamma_H$ must be updated to reflect that observation. Can this be done quickly and efficiently?

The answer is definitely negative: example \ref{example:compass0} and figure \ref{figure:compass05} demonstrate the fact that $\Gamma_H$ can change immensely as a result of seemingly minor adjustments of $H$ (e.g., in the example, if $\epsilon$ is very close to $\pi/4$, a very small change in $\epsilon$ will cause a cycle to collapse into a tree or vice-versa). Given that $\Gamma_H$ may resemble practically any big graph on the large scale, the problem of structural updating for $\Gamma_H$ becomes intractable very quickly.

Despite the failure of our first attempt, we will construct a workable model using an almost identical skeleton of ideas. After all, $\Gamma_H$ does have some very desirable traits one would be happy to retain:
\begin{itemize}
	\item[-] $\Gamma_H$ encodes rudimentary logic (negation, implication), 
	\item[-] it seems possible to synchronize updating content with structural updating (e.g., erase vertices representing visible states of little interest),
	\item[-] a {\it learning goal} for $\hsm{O}$ can be defined: have $\Gamma_H$ grow to be big/detailed enough, with its content eventually close enough to objective values, in order to guarantee the success of $\hsm{O}$.
\end{itemize}

\subsection{Results} 
The model of memory we propose in this paper, has an underlying category whose objects are duals, in some sense, of graphs belonging to the well studied family of {\it median graphs}. The interested reader should see \cite{[Rol]} for an extended bibliography on the subject and a detailed treatment of duality theory for median algebras, and median graphs in particular. A very good treatment of median graphs and their duality theory is given in \cite{[Nica]}. We give a self-contained exposition of the relevant notions and results in section \ref{section:basic model}, though we do focus on the current application and omit proofs of results from the literature.

One remarkable feature of our model is that the structure of a median graph itself provides our observers with a rudimentary sense of logic. We will discuss this formal idea of `common sense' as we progress through the next section.

Searching our databases for content and updating content is discussed in section \ref{section:implementation}. We explain how searching our database structures essentially coincides with content updating and discuss some of the implications regarding learning. A possible implementation of the model as a network of neuron-like elements is offered as well, and we discuss its computational efficiency.

Section \ref{section:categorification} is dedicated to structural updating and a discussion of how the phenomena of learning, language formation, forgetting and understanding are realized in the model.

As the exposition evolves, we periodically pause to look at how various natural phenomena are accounted for by our model. Right now we can state, with some satisfaction, that our model explains an overwhelming majority of the phenomena we had already listed as curious aspects of the human thought process and human memory.

The last section discusses weaknesses of the model and possible ways to get rid of them -- a topic for future research.

\subsection*{Acknowledgements} The author is deeply indebted to Michael Jablonski and Vera Tonic for proof-reading the text and for numerous useful comments on both content and form. Many thanks to Lucas Sabalka for the idea of replacing computationally cumbersome examples of spaces with walls with the compass example (which is yet to reappear in our narrative); to Kre\v{s}imir Josi\v{c} for commentary on the material of section \ref{section:implementation} and ongoing discourse. The author is a newcomer to the field, and must confess limited knowledge of existing literature. I am grateful in advance for any comments and criticisms, and will happily acknowledge credit wherever credit is due.

\section{Modeling memory using poc-sets and median graphs}\label{section:basic model}
\subsection{Poc-sets.} We consider an abstract version of a system of questions, due to Roller \cite{[Rol]}:
\begin{defn}[poc-set]\label{defn:poc-set} A poc-set $(P,\leq,\ast)$ is a partially-ordered set $(P,\leq)$ with a minimum (denoted by $0$), endowed with an order-reversing involution $a\mapsto a^\ast$ such that $a\leq a^\ast$ implies $a=0$ for all $a\in P$.
The maximum $0^\ast\in P$ is denoted by $1$; the elements $0,1\in P$ are said to be \emph{trivial}; all other $a\in P$ are \emph{proper}.
\end{defn}
\begin{example}[standard poc-sets] The Borel $\sigma$-algebra $\hsm{B}$ carries a natural poc-set structure $(\hsm{B},\subseteq,a\mapsto a^c)$. If $P$ is a poc-set and $0\in Q\subseteq P$ satisfies $Q^\ast=Q$ then $Q$ is a (sub) poc-set (of $P$).
\end{example}
For any proper $a,b\in P$, it is easy to see that only one of the following may hold:
\begin{equation}\label{eqn:nesting relations}
	a\leq b\,,\quad a\leq b^\ast\,,\quad a^\ast\leq b\,,\quad a^\ast\leq b^\ast.
\end{equation}
\begin{defn}[nesting, transversality] A pair $a,b$ of elements in a poc-set $P$ is said to be \emph{nested}, if any one of the relations in \ref{eqn:nesting relations} holds. Otherwise, the pair $a,b$ is said to be \emph{transverse} (denoted $a\pitchfork b$). A subset $S\subseteq P$ is said to be nested (resp. transverse), if the elements of $S$ are pairwise nested (resp. transverse).
\end{defn}
A means for relating poc-sets to each other will be required:
\begin{defn}[morphisms]\label{defn:poc-morphism} A function $f:P\to Q$ between poc-sets is a \emph{morphism} if $f(0)=0$, $f$ is order-preserving and $f(a^\ast)=f(a)^\ast$ for all $a\in P$.
\end{defn}
We were previously considering the possibility of modeling the memory of an observer $\OO$ by a sub poc-set of $\hsm{B}$, but it works better to use a pair $(P,f)$, with $P$ an abstract poc-set and $f:P\to\hsm{B}$ a morphism. We shall presently see that the abstract object $P$ gives rise to a graph $\Gamma(P)$ such that: (a) $P$ is reconstructible from $\Gamma(P)$ and (b) the graph $\Gamma(P,f)=\Gamma_{f(P)}$ of visible states determined by $f(P)$ (see discussion in sub-section \ref{subsection:stating the problem}) is canonically embedded in $\Gamma(P)$. Thus, if we choose $\Gamma(P)$ to represent the memory of $\OO$ while $f$ is viewed as an interpretation of $P$ in the reality presented by $X$, then the above property of $\Gamma(P)$ implies that no updating of the structure of the memory graph $\Gamma(P)$ is required so long as the implication relations among the questions available to $\OO$ (that is -- the structure of $P$) remain unchanged. Not taking other aspects of the modeling problem into account, properties (a) and (b) of $\Gamma(P)$ should be viewed as the main argument in favor of preferring $\Gamma(P)$ over $\Gamma(P,f)$: while containing complete information about $P$ (and hence not being `too big'), $\Gamma(P)$ has `sufficient space' to accommodate {\it any} possible interpretation of $P$ in $X$.\\

Perhaps it is time again for a concrete example.
\begin{example}[Compass, part 2]\label{example:compass1} We return to example \ref{example:compass0}. Recall the sets $N,S,W,E$ defined there as arcs of length $2\epsilon$, $\epsilon\in(0,\pi/2)$, on the unit circle -- see figure \ref{fig:compass1} and example \ref{example:compass0}. This time, define a formal poc-set $P$ as follows: 
\[P=\left\{0,1,n,s,w,e,n^\ast,s^\ast,w^\ast,e^\ast\right\}\,,\] and we set $f(n)=N$, $f(s)=S$ and so on, together with the relations following from the requirement that $f$ be a morphism.

\begin{figure}[tb]
	\centering{\includegraphics[width=0.75\textwidth]{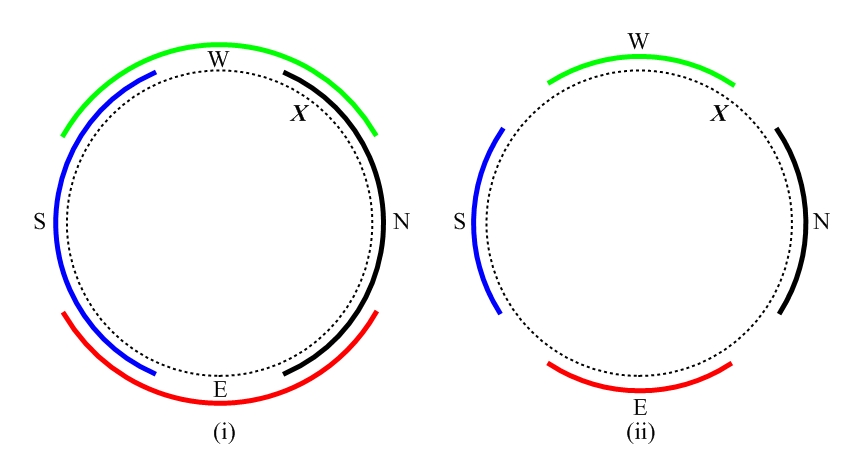}}
	\caption{\protect\scriptsize Observing a compass (1). Two possible realizations of the same abstract poc-set structure.\protect\normalsize
		\label{fig:compass1}}	
\end{figure}

We will compare two situations, see figure \ref{fig:compass1}: (i) the one where $\epsilon\in(\pi/4,\pi/2)$ with (ii) the one where $\epsilon\in(0,\pi/4)$.

Observe that $\epsilon<\pi/2$ implies that a positive answer to $n$ excludes a positive answer to $s$ and vice versa. The same is true about the pair $\{e,w\}$. We have chosen $\epsilon$ in this way in order to enable a choice of the poc-set structure on $P$ having the relations $n\leq s^\ast$ and $w\leq e^\ast$ and all conclusions thereof. Both in this example and in its continuation (example \ref{example:compass2}) $P$ will be chosen to have these relations (and their consequences). To be sure, we list the relations on $P$:
\begin{eqnarray*}
	n<s^\ast, & n\pitchfork e, & n\pitchfork w,\\
	s<n^\ast, & s\pitchfork e, & s\pitchfork w,\\
	e<w^\ast, & w<e^\ast\,.
\end{eqnarray*}
Note that if we had wanted to allow similar realizations of $P$ with $\epsilon>\pi/2$, we would have been forced to change all the above relations into transversality relations, in which situation the poc-set $P$ on its own would not have held {\it any} information about the observed system.

When $\epsilon<\pi/4$ (see (ii) in the figure), the sets $N,S,E,W$ are pairwise complementary. This means that the poc-set $f(P)$ has the relations $N<E^c$ and $N<W^c$ although neither $n<e^\ast$ nor $n<w^\ast$ hold in $P$.

For $\epsilon>\pi/4$ though (see (i) in the figure), all the relations holding in $f(P)$ are accounted for already in $P$. In example \ref{example:compass2} we will see how this difference between the situations is visualized at the level of the corresponding graphs. 
\end{example}

\subsection{Sageev-Roller duality.}
\begin{figure}[h]
	\centering{\includegraphics[width=\textwidth]{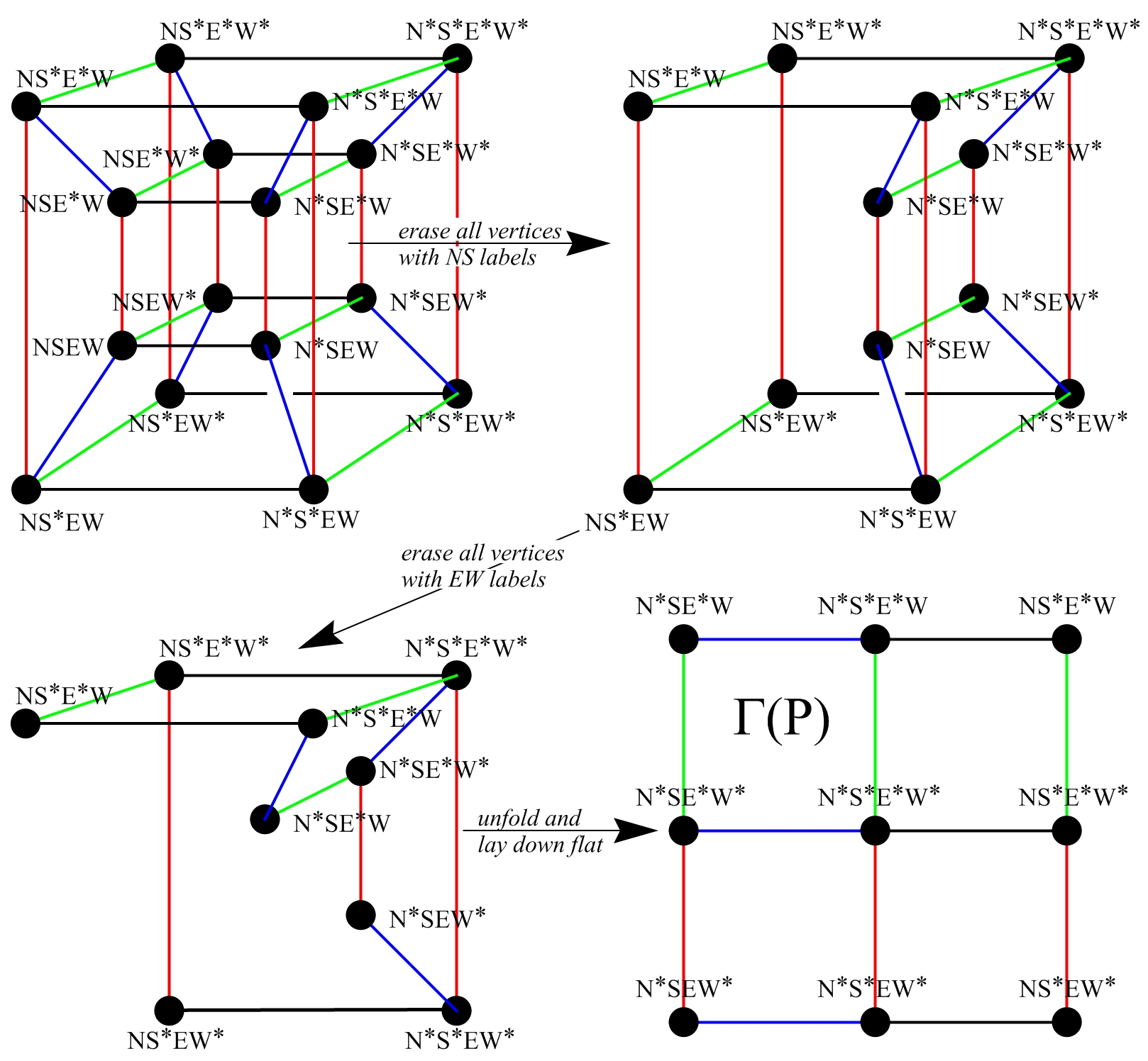}}
	\caption{\protect\scriptsize Observing a compass (2). The graph $\Gamma(P)$ for our model of an observer of a compass as constructed in example \ref{example:compass1}\protect\normalsize
		\label{fig:compass2}}
\end{figure}

\begin{figure}[tb]
	\centering{\includegraphics[width=0.5\textwidth]{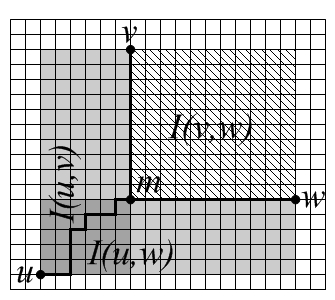}}
	\caption{\protect\scriptsize The Integer Grid: example of computing a median in a median graph.\protect\normalsize
	\label{fig:median in grid}}
\end{figure}

The construction of $\Gamma(P)$ -- the graph {\it dual} to $P$ -- given in this paragraph is originally due to Sageev \cite{[Sa]} in a special case. The presentation we have chosen and the discussion of dual maps is due to Roller \cite{[Rol]}. However, we have chosen to alter Roller's original terminology to better fit the intended application.
\begin{defn}[coherence, vertices] Let $(P,\leq,\ast)$ be a poc-set. A subset $\alpha\subset P$ is said to be coherent if $a\leq b^\ast$ holds for \emph{no} $a,b\in\alpha$. A maximal coherent family will be called a \emph{vertex of the graph $\Gamma(P)$}.
\end{defn}
It is easy to see that a coherent family $\alpha\subset P$ lies in $V\Gamma(P)$ iff $\alpha$ is a $\ast$-selection on $P$ (meaning that for all $a\in P$, either $a\in\alpha$ or $a^\ast\in\alpha$, but not both).
\begin{example} Let $\mathbf{2}=\{0,1\}$ be the {\it trivial poc-set}, with the obvious relation $0<1$. Show that the assignment $f\mapsto f^{-1}(1)$ from the set $\mathrm{Hom}(P,\mathbf{2})$ of all poc-morphisms of $f:P\to\mathbf{2}$ to the set $V\Gamma(P)$ is a bijection.
\end{example}
\begin{defn}[edges] Let $(P,\leq,\ast)$ be a poc-set and $u,v\in V\Gamma(P)$. We set $\{u,v\}\in E\Gamma(P)$ iff $u\vartriangle v=\{a,a^\ast\}$ for some $a\in P$. Here $u\vartriangle v$ denotes the symmetric difference $(u\minus v)\cup(v\minus u)$.
\end{defn}
More generally, for $u,v\in V\Gamma(P)$ one has $\card{u\vartriangle v}=2\card{u\cap v^\ast}$, which implies that the expression $\Delta(u,v)=\frac{1}{2}\card{u\vartriangle v}$ -- the number of questions separating $u$ from $v$ -- is a distance function on $V\Gamma(P)$. Moreover, $\Delta(u,v)=1$ iff $u$ and $v$ are joined by an edge in $\Gamma(P)$.
\begin{example}[Compass, part 3]\label{example:compass2} 
We would like to go back to the poc-set \[P=\left\{0,1,n,s,w,e,n^\ast,s^\ast,w^\ast,e^\ast\right\}\,\] defined in example \ref{example:compass1} and endowed with the structure
\begin{eqnarray*}
	n<s^\ast, & n\pitchfork e, & n\pitchfork w,\\
	s<n^\ast, & s\pitchfork e, & s\pitchfork w,\\
	e<w^\ast, & w<e^\ast\,.
\end{eqnarray*}

In order to visualize $V\Gamma(P)$, consider $\Gamma(P)$ as a full simple subgraph of the cube with vertex set $2^P$ (where each vertex corresponds to a choice of one answer for every question, ignoring coherence issues): $\Gamma(P)$ is the result of erasing all vertices of $2^P$ turning out to be incoherent, together with their adjacent edges -- see figure \ref{fig:compass2}. 
\end{example}
\begin{example}[The $n$-cube]\label{example:cube} Let $T$ be a set of $n$ distinct symbols, and let $P$ be the poc-set generated by $T$ with no relations, that is: $P=\{0,0^\ast\}\cup T\cup T^\ast$, where $T^\ast$ is the set of symbols of the form $t^\ast$ such that $T^\ast\cap T=\varnothing$, and no two distinct proper elements of $P$ are comparable.

Then any $\ast$-selection on $P$ is a maximal coherent subfamily, and we conclude that $\Gamma(P)$ is the $1$-dimensional skeleton of the $n$-dimensional cube.
\end{example}
The next example is based on the following recommended easy exercise:
\begin{example} We offer the following exercise to the reader. Suppose $P$ is a finite poc-set and let $a\in P$. Prove that the following are equivalent:
\begin{enumerate}
	\item $a$ is nested with every element of $P$,
	\item $\Gamma(P)$ has one and only one edge $\{u,v\}$ satisfying $u\vartriangle v=\{a,a^\ast\}$, and this edge is a cut-edge.
\end{enumerate}
\end{example}
\begin{example}[Trees and $n$-pompoms]\label{example:pompom} The above exercise shows that if $P$ is nested then $\Gamma(P)$ is a tree. The converse is known to be true as well (exercise). A special case that is very easy to compute is constructed as follows: take $A$ to be a set of $n$ distinct symbols and let $P=\{0,0^\ast\}\cup A\cup A^\ast$ where $A$ and $A^\ast$ are disjoint and subject to the relations $a_i^\ast<a_j$ for all $1\leq i<j\leq n$. This immediately implies there are only two kinds of vertices: the vertex $v_0=\{0^\ast\}\cup A$ and the vertices $v_i=\{0^\ast,a_i^\ast\}\cup\left(A\minus\{a_i\}\right)$. Then $\Gamma(P)$ is the tree with $(n+1)$ vertices and $n$ leaves $v_1,\ldots,v_n$. We will refer to this structure as the $n$-pompom.
\end{example}
For any connected graph $\Gamma$ and $u,v\in V\Gamma$ the \emph{interval between $u$ and $v$} is defined as
\begin{equation}
	I(u,v)=\left\{w\in V\Gamma\,\left|\;d_\Gamma(u,w)+d_\Gamma(w,v)=d_\Gamma(u,v)\right.\right\}\,,
\end{equation}
where $d_\Gamma$ denotes the path distance on $\Gamma$. $I(u,v)$ is clearly the union of the vertex sets of shortest paths (geodesics) from $u$ to $v$.
\begin{defn}[median graph]\label{defn:median graph} A connected graph $\Gamma$ is a median graph if, for all $u,v,w\in V\Gamma$ the intersection $I(u,v)\cap I(v,w)\cap I(u,w)$ contains precisely one vertex (denoted $\med(u,v,w)$).
\end{defn}
An example visualizing what is going on is the standard rectangular integer grid:
\begin{example}[The Grid] Let $X$ be the set of points $(x,y)\in\fat{R}^2$ satisfying $x\in[0,m+1]$ and $y\in[0,n+1]$ with $m,n$ being positive integers. Let
\begin{equation*}
	h_s=\left\{(x,y)\,\Big|\,y<s+\frac{1}{2}\right\}\,,\qquad
	v_t=\left\{(x,y)\,\Big|\,x<t+\frac{1}{2}\right\}\,,
\end{equation*}
where $s\in\{1,\ldots,n\}$ and $t\in\{1,\ldots,m\}$. Then the set
\begin{equation*}
	P=\{\varnothing,X\}\cup\left\{h_s,h_s^c\right\}_{s=1}^n\cup\left\{v_t,v_t^c\right\}_{t=1}^m
\end{equation*}
is a sub poc-set of $\hsm{B}$ (the Borel $\sigma$-algebra on $X$) with respect to inclusion $(\subseteq)$ and complementation $(A\mapsto A^c)$. It is easy to see that $h_s\pitchfork v_t$ for all $s,t$, while $h_s\subset h_{s+1}$ and $v_t\subset v_{t+1}$ for all relevant $s$ and $t$. As a result, any subset of $P$ is consistent if and only if it is coherent, and the visible states of $X$ with respect to $P$ (letting $P$ be realized by the inclusion map) are in one-to-one correspondence with the vertices of $\Gamma(P)$. Observe that the visible states are in one-to-one correspondence with the integer points of $X$, so it makes sense to use these points as representatives, joining two such points by a straight line segment if and only if the corresponding vertices of $\Gamma(P)$ are joined by an edge. Figure \ref{fig:median in grid} demonstrates the computation of a median of three vertices of $\Gamma(P)$, drawn over a diagram of $\Gamma(P)$ realized in this way in the plane.
\end{example}
A good exercise for the interested reader will be to prove the following --
\begin{lemma}[median operation] If $(P,\leq,\ast)$ is a finite poc-set, then $\Gamma(P)$ is a median graph, and
$\med(u,v,w)=(u\cap v)\cup(v\cap w)\cup(u\cap w)$ for all $u,v,w\in V\Gamma(P)$.
\end{lemma}
\begin{thm}[`Sageev-Roller duality', \cite{[Rol]}] If $P$ is a finite poc-set then $\Gamma(P)$ is a connected median graph and $d_{\Gamma(P)}$ coincides with $\Delta$. Furthermore, every finite median graph arises in this way.
\end{thm}
By this theorem, the construction of an arbitrary `memory graph' makes it automatically a median graph. Also, every median graph can be thought of as a `memory graph' when provided with a realization map relating it to a state space of an observed system. We conclude that the property of being a median graph can be identified as a one of the ``guiding principles'' mentioned in the introduction -- a principle restricting the structure of such graphs. 

We shall now proceed to demonstrate the applications of this idea using the study of morphisms of median graphs (maps between median graphs preserving the median structure).\\

An additional aspect of the above duality is that morphisms of poc-sets translate into median morphisms of median graphs and vice versa. If $A$ and $B$ are median graphs, then a morphism from $A$ to $B$ is a function $VA\to VB$ preserving medians.
\begin{remark} We do not require a morphism of median graphs to preserve the adjacency relation. In fact, imposing this additional requirement proves to be overly restrictive for our purposes. 
\end{remark}
If $f:P\to Q$ is a morphism of finite poc-sets, then a \emph{dual morphism} of median graphs $f^\circ:\Gamma(Q)\to\Gamma(P)$ is defined by $f^\circ(v)=f\inv (v)$ for $v\in V\Gamma(Q)$. Some properties of this construction are:
\begin{prop}[functorial properties, see \cite{[Rol]}] If $f:P\to Q$, $g:Q\to R$ are morphisms of finite poc-sets then:
\begin{enumerate}
	\item $(g\circ f)^\circ=f^\circ\circ g^\circ$;
	\item $f$ is surjective if and only if $f^\circ$ is injective;
	\item $f$ is an embedding if and only if $f^\circ$ is surjective.
\end{enumerate}
(by an embedding we mean an isomorphism onto the image)
\end{prop}
\begin{figure}[tb]
	\centering{\includegraphics[width=0.75\textwidth]{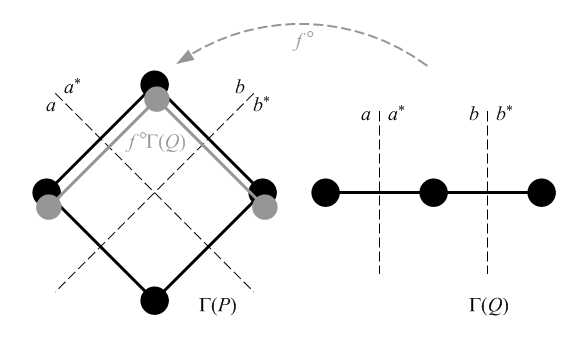}}
	\caption{\protect\scriptsize Visualizing the dual map of $f$ from example \ref{example:injective but not an embedding}.\protect\normalsize
	\label{fig:injective but not an embedding}}
\end{figure}
\begin{example}[see fig. \ref{fig:injective but not an embedding}]\label{example:injective but not an embedding}
Let $P$ and $Q$ be poc-sets with $4$ proper elements each -- $a,a^\ast,b,b^\ast$ -- such that $a\pitchfork b$ in $P$ but $a<b$ in $Q$. In this case, the set-theoretic identity map $f=id:P\to Q$ is a morphism, while its inverse is not. $\Gamma(P)$ is a $4$-cycle, while $\Gamma(Q)$ is a path of length $2$ (having $3$ vertices). While $f$ is bijective, it is not an embedding: though $f\inv $ is well-defined, it is not a morphism of poc-sets.
\end{example}

\subsection{Consistent families and weights: our model of an observer}
We now return to the idea of a morphism $f:P\to\hsm{B}$ representing the memory of an observer at a fixed moment in time. We will henceforth refer to such $f$ as a \emph{representation} of $P$.

For every $x\in X$, consider the maximal coherent subfamily of $\hsm{B}$:
\begin{equation}
	\pi_x=\left\{A\in\hsm{B}\,\left|\,x\in A\right.\right\}\,.
\end{equation}
One can apply $f^\circ$ to $\pi_x$ to obtain the maximal coherent family
\begin{equation}
	\pi_{P,f}(x)=\left\{a\in P\,\left|\,x\in f(a)\right.\right\}\,\in V\Gamma(P).
\end{equation}
We now have a map $\pi_{P,f}:X\to V\Gamma(P)$ selecting precisely those vertices in $\Gamma(P)$ corresponding to visible states of $X$ relative to $f(P)$. Since $\pi_{P,f}$ is constant on every visible state, $\pi_{P,f}$ induces an injective map from $V\Gamma(P,f)$ into $V\Gamma(P)$. From the definition of edges in both graphs it is clear that this injection is an embedding of graphs, which we denote by $\bar\pi_{P,f}:\Gamma(P,f)\to\Gamma(P)$.
\begin{defn}[consistent families] A set $u\subseteq P$ is said to be consistent relative to a representation $f$, if there is a point $x\in X$ such that $f(u)\subseteq\pi_x$.
\end{defn}
In particular, a vertex $u$ of $\Gamma(P)$ is consistent iff it lies in the image of $\Gamma(P,f)$ under $\bar\pi_{P,f}$. We illustrate this on the compass example (examples \ref{example:compass1} and \ref{example:compass2}):
\begin{figure}[tb]
	\centering{\includegraphics[width=\textwidth]{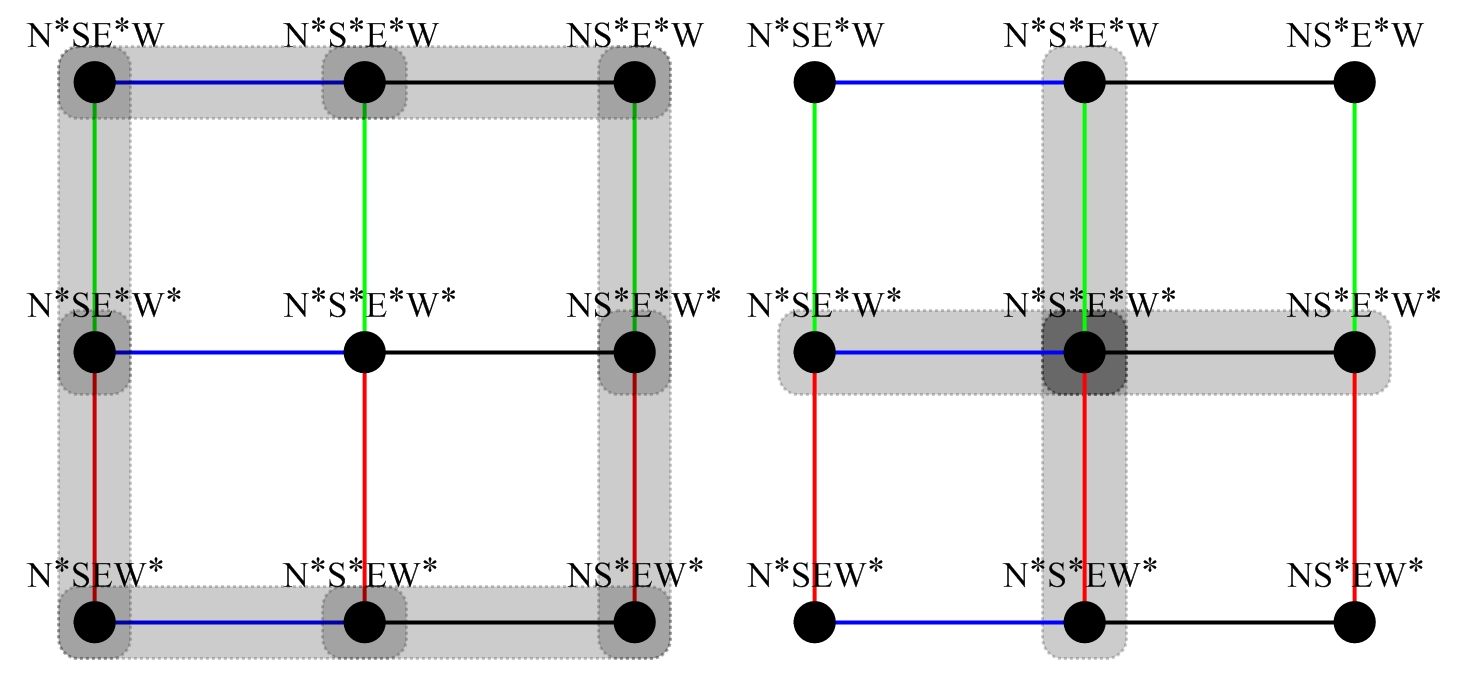}}
	\caption{\protect\scriptsize Observing a compass (2). The two realizations from example \ref{example:compass1} induce {\it very} different embeddings of $\Gamma(P,f)$ in $\Gamma(P)$: for $\epsilon>\pi/4$ (left) and for $\epsilon<\pi/4$ (right). Edges of $\Gamma(P)$ belonging to $\Gamma(P,f)$ are marked by transparent boxes drawn on top of them.\protect\normalsize
	\label{fig:compass3}}
\end{figure}
\begin{example}[Compass, part 4]\label{example:compass3} Figure \ref{fig:compass3} shows $\Gamma(P,f)$ embedded in $\Gamma(P)$. It is important to observe that the central vertex is inconsistent for $\epsilon>\pi/4$, whereas the same vertex becomes consistent upon reducing $\epsilon$ to a value less than $\pi/4$, with the corner vertices becoming inconsistent in this situation. 

Looking at the picture leaves one amused by the fact that, somehow, smaller precision in answering the questions $n,s,w$ and $e$ (corresponding to $\epsilon>\pi/4$) contributes to a better discrete model of $X$ (recall $X$ was a circle) than the one obtained from more precise answers. More than that, we consider this example as clear proof of our claim that separating the logical structure $P$ from its realization (resulting in making the distinction between $\Gamma(P)$ and $\Gamma(P,f)$) constitutes a significant improvement over the simplistic model described in the introduction: indeed, $\Gamma(P)$ in this case is spacious enough to include faithful representations of both types of realization that we had discussed, while the separate realizations seem completely incompatible (compare the two versions of $\Gamma(P,f)$ in figure \ref{fig:compass3} again).
\end{example}

Now we are ready to formally define a model of an observer:
\begin{defn}\label{defn:observer} Let $(X,\hsm{B},\mu)$ be a probability space. An {\it observer} of $X$ is a quadruple $\OO=(P,f,p,\epsilon)$, where $P$ is a finite poc-set, $f:P\to\hsm{B}$ is a morphism of poc-sets, $p:V\Gamma(P)\to[0,\infty)$ is a function on the vertices of $\Gamma(P)$ and $\epsilon$ is a subset of $P$.
\end{defn}
The additional data elements -- $p$ and $\epsilon$ -- are new to our discussion. Ethological\footnote[1]{Ethology -- a field of Biology, studying animal behaviour.} considerations seem to imply that memory is subject to prioritization: different events are considered relevant to different extents depending on the type of conflict they create between the individual and the environment. The types of conflict can be classified, and the subjective perception of the intensity of an interaction between the individual an the environment may be measured by observing the change in the level of physiological stress resulting from the interaction \cite{[Guralnik-Hierarch]}. In this context, we want $p(v)$ to represent an estimation that $\OO$ has regarding the relevance of the event $\bigcap_{a\in v}f(a)$, for every $v\in V\Gamma(P)$, and $\epsilon$ to represent the conjecture $\OO$ has regarding the current state of the universe. Ideally, $\epsilon$ will be a vertex of $\Gamma(P)$. Less ideally, $\epsilon$ is a coherent family. One should note, however, that humans holding a completely coherent set of convictions are rarely to be found.\\

We will need some new notation:
\begin{defn}[convexity, convex hull] 
In a graph $\Gamma$, one says that a subset $W$ of $V\Gamma$ is {\it convex}, if $I(u,v)$ is contained in $W$ for all $u,v\in W$. The {\it convex hull} $conv(W)$ of $W$ is defined to be the intersection of all convex subsets of $V\Gamma$ containing $W$. 
\end{defn}
Clearly, $conv(W)$ is the smallest convex subset of $V\Gamma$ containing $W$.
\begin{defn} Let $P$ be a poc-set. For $a\in P$ and $A\subseteq P$ we denote:
\begin{equation}
	V(a)=\left\{u\in V\Gamma(P)\left|a\in u\right.\right\}\,,\qquad V(A)=\bigcap_{a\in A}V(a)\,.
\end{equation}
\end{defn}
Observe that $V(a^\ast)=V\Gamma(P)\minus V(a)$ for all $a\in P$. Thus, the family $\left\{V(a)\right\}_{a\in P}$ forms a poc-set with respect to inclusion and complementation. In fact, it is easy to see that this poc-set is isomorphic to $P$ via the map $a\mapsto V(a)$. For the case of a median graph, when one can safely write $\Gamma=\Gamma(P)$, it turns out that $conv(W)=\bigcap_{W\subset V(a)}V(a)$. Thus, $W\subset V\Gamma(P)$ is convex if and only if it equals the intersection of a family of halfspaces of $V\Gamma(P)$.\\

Let us now return to the discussion of the relationship between a living observer and our notion of an observer $\OO=(P,f,p,\epsilon)$. We first focus our attention on the function $p$. As stated above, $p$ should be thought of as a measure of the relevance of a vertex $u\in V\Gamma(P)$ in the observer's eyes: the event $u$ is likely to be disregarded by the observer if $p(u)$ is negligible, while an event of the form $V(A)$ will attract more of the observer's attention the higher the cumulative value $p(V(A))=\sum_{u\in V(A)}p(u)$. Thus, from an information-theoretic point of view, every event $F\subset V\Gamma(P)$ has a probability attached to it:
\begin{equation}
	\PR{\OO}{F}=\frac{p(F)}{p\left(V\Gamma(P)\right)}
\end{equation} 
This should not be confused with the function taken on by the set $\epsilon$ corresponding to the observer's conjecture of the `current state of affairs': $\PR{\OO}{V(\epsilon)}$ being small means our observer is aware of no sources of stress right now; the same quantity being large {\it may} provide a motivation for our observer to take action to change $\epsilon$ in the direction of lowering the stressfulness of the situation. 

However, one should not imagine living organisms as trying to solve some kind optimization problem (with the objective being to minimize stress), but, rather, as trying to solve an equilibrium problem. Ethological studies \cite{[Guralnik-Hierarch]} show that minimization of stress could not be regarded as a plausible goal for {\it every} observer. Our information-theoretic interpretation of $p$ is a convenient tool for describing this phenomenon: an observer cannot be attracted to the idea of pushing $X$ into a state which this same observer sees as uninteresting.\\

Here are some possible combinations of $f$, $\epsilon$ and $p$ to contemplate:
\begin{example}[objectivity] Given $\OO=(P,f,p,\epsilon)$, choose $p(u)=\mu\left(\bigcap_{a\in u}f(a)\right)$ and $\epsilon=\pi_x$ where $x$ is the actual current state of the system. The resulting observer is, in some sense, objective: his perception of events is precise ($\epsilon$ is a maximal coherent family), up to date ($V(\epsilon)$ includes the current state) and unbiased ($\PR{\OO}(\cdot)$ gives the correct probabilities of events).
\end{example}
\begin{example}[misperception] Once again, suppose $P$ and $f$ are given, and $x$ is the current state of the universe. Consider some possible situations for an observer $\OO=(P,f,p,\epsilon)$:
\begin{description}
	\item[$\epsilon\subsetneq\pi_{P,f}(x)$] This means that $\OO$ has a question allowing it to improve its perception of the current state of the universe.
	\item[$\epsilon\not\subseteq\pi_{P,f}(x)$] This means that $\OO$ has a question which -- had it been asked -- would have proved the observer's view of the universe to be inconsistent with the current state, forcing the observer to alter its structure as an observer.
	\item[$p(u)=0$ with $u$ consistent] In this situation $\OO$ regards an existing state of the universe as impossible/negligible.
	\item[$p(u)>0$ with $u$ inconsistent] Here an impossible event ($u$) may attract the attention of $\OO$, who considers this event as disturbing (and hence probable).
\end{description}
The above situations demonstrate the flexibility of our model. They also motivate looking for a way to measure how far an observer $\OO$ is from being objective.
\end{example}

\subsection{$\Gamma(P)$ as a representation of `common sense'}\label{subsection:common sense} The preceding examples are concerned with the possible relations between the content components $p$ and $\epsilon$ of an observer $\hsm{O}=(P,f,p,\epsilon)$ and its representation map. Now we would like to focus on the structural component $P$ and its relation to so-called `common sense'.\\

By common sense we do not mean ideas or rules common in society: `cannibalism is bad' is not an inborn common notion, but a fundamental and non-trivial cornerstone of modern society. By common sense we mean a certain shared notion of rudimentary logic that is inherent to all our actions. In part, this notion is embodied in our model through the assumption that memory is structured by a poc-set, and realized by a poc-morphism. Consider an observer $\OO$ as in definition \ref{defn:observer}. Both the structure of $P$ and that of $\Gamma(P)$ carry information about implication relations among events in $X$ as those are perceived by $\OO$. This is due to the equivalence $a<b\IFF V(a)\subset V(b)$ holding for all $a,b\in P$. Together with the fact that $V(a^\ast)=V(a)^c$, this provides an automated tool $\OO$ can use for `sub-conscious' reasoning.

Thus, all our observers share the way in which the content of their memory is ordered, and this fact is bound to affect the ways in which two observers sharing the same environment (or territory) synchronize their actions.

When is it that we are able to demonstrate to others that we understand something well? The pedagogical answer has always been that good understanding is defined as a state in which discussing the subject matter adequately from a logical standpoint does not require a lot of conscious effort.\\

However, logic as we understand it involves the ability to operate with combined observations. While implication and negation are inherent to observers through the poc-set structure of the `atomic' statements, conjunctions (and, dually, disjunctions) are not taken into account directly by the poc-set structure.\\

To demonstrate this, fix a positive integer $n$ and consider two hypothetical observers: 
\begin{itemize}
	\item[-] $\OO_1$ with poc-set component $P_1$ such that $\Gamma(P_1)$ is the $n$-cube (see example \ref{example:cube}), and an excitation function $p_1$ assigning a unit value to every vertex;
	\item[-] $\OO_2$ with poc-set structure $P_2$ such that $\Gamma(P_2$ is the $2^n$-pompom (see example \ref{example:pompom}) with center $v_0$, and with $p_2$ assigning unit excitation values to all the leaves of $\Gamma(P_2)$ and null excitation to $v_0$.	
\end{itemize}
Furthermore, suppose $f_1$ realizes $\OO_1$ so that the partition $\hsm{P}(f_1(P_1))$ induced on $X$ is uniform with respect to $\mu$ (in particular, $\OO_1$ is objective). We do not want $\OO_2$ to be at an unfair disadvantage, so assume the partition $\hsm{P}(f_2(P_2))$ coincides with $\hsm{P}(f_1(P_1))$ so that $v_0$ is $f_2$-inconsistent. Thus, $\OO_2$ is objective as well, and the only way in which the two observers differ is the combinatorics each of them uses to model the same partition of $X$.\\

We will now ask the standard question an information theorist asks in situations like this: what is the expected minimum number of observations that each observer needs to make about the current state in order to identify the vertex in its memory corresponding to this state of $X$? Then answers are clearly very different: $n$ for $\OO_1$ and $2^n-1$ for $\OO_2$.\\

A natural way for $\OO_2$ to come closer to the optimal position of $\OO_1$ is to widen the supply of direct observations $\OO_2$ is able to make. If $\OO_2$ had a mechanism for expanding $P_2$ by adding conjunctions of elements from $P_2$ to it, the updated version of $\OO_2$ would be able to perceive $X$ with lower entropy.\\

On the other hand, it is not reasonable for $\OO_1$ to maintain the strategy of keeping $P_1$ completely transverse (except for trivial nesting relations) over a long time: as the number of available observation tools increases, $\Gamma(P_1)$ will keep growing exponentially, without ever encoding {\it any} of the recorded information in its combinatorial structure.\\

The inevitable conclusion is that an approach balancing content and structure should exist, and we expect it to become all the more efficient as the particular implementation of the memory structure comes to possess tools allowing the observer to refine its observations by combining them at will.

\section{Possible Implementation and Information Retrieval}\label{section:implementation}
\subsection{The basic searching problem.} The goal of this section will be to define the basic searching problem for an observer $\OO=(P,f,p,\epsilon)$ and to discuss the main features of an implementation that one may regard as efficient.

Deferring updating tasks which involve altering the structure of $P$ or the excitation function $p$, one is left with the task of efficiently updating $\epsilon$ -- the observer's conjecture about the current state of events -- in response to an incoming observation. 

The basic search/retrieval/update problem may be formulated as follows. Suppose an observer modeled by $\OO=(P,f,p,\epsilon)$ has just made the observation $a\in P$. Then there is a need to (1) decide whether $\epsilon\cup\{a\}$ is consistent, and then (2) replace $\epsilon$ by $\epsilon\cup\{a\}$ in case it is, or (3) replace $\epsilon$ by a new description of the perceived current state that is as consistent as possible with $\epsilon$ and the new observation.

This is the point when any discussion of efficiency will depend on the specific implementation, that is: on {\it how} the memory structure is realized and maintained. In order to facilitate this discussion, let us consider the different components of an observer $\OO$ from a more practical point of view:
\begin{itemize}
	\item Our initial assumption was that observing (and recording observations about) a system with a given state space $X$ may be thought of as maintaining a database whose purpose is to mimic the geometry and topology of $X$, as those are revealed through asking binary questions about $X$. Such a question can be very basic, e.g. ``Is neuron number 1234567 firing right now?''.
	\item The function $f$ in the definition of an observer provides the connection to reality: $f$ essentially represents the sensors used by the observer for watching the evolution of the observed system, e.g. neuron 1234567 fires if and only if a certain patch of light receptors on our retina gets hit by a sufficient number of photons. 
	\item Both $\Gamma(P)$ and $P$ are maintained (stored in memory) as directed graphs, with the vertices of $\Gamma(P)$ labeled by their values under the function $p$ and the edges of $\Gamma(P)$ labeled by elements of $P$.
	\item Finally, recording $\epsilon$ is no more than a labeling on the vertices of the graph representing $P$. This graph is nothing but a Hasse diagram with additional edges labeled by $(\ast)$ to join every $a\in P$ to $a^\ast\in P$. The labeling corresponding to $\epsilon$ works as follows: a vertex $a\in P$ is `ON' if and only if $a\in\epsilon$. 
\end{itemize}

\subsection{Idealized searching.}\label{subsection:idealized searching} Let us now suppose that an implementation of $P$ is equipped with the following {\it idealized} features in addition to the graph structure we had just described:
\begin{description}
	\item[Propagating Excitation] Every node of $P$ (we identify elements of $P$ with the corresponding nodes) has an excited state and a non-excited state. If $a\in P$ is excited, then so is every $b\in P$ with $b>a$.
	\item[Contradiction Detection] If $a\in P$ is excited, and $a^\ast$ is `ON', then a flag is raised and the detector outputs $a$.
\end{description}
The first feature is motivated by neurons, but very far from being a precise copy of the same mechanism. Upon observing the algorithmic implications of the idealized feature we will de-idealize it and discuss the consequences.\\

Suppose an observer modeled by $\OO$ makes an observation $a\in P$. The situation then requires a reaction on the part of the observer with the aim to keep $\OO$ up to date: (1) if $a\in\epsilon$ then $\epsilon$ is not changed; if $a\not\in\epsilon$, then either (2) $\epsilon\cup\{a\}$ is coherent and $\epsilon$ will be replaced by $\epsilon\cup\{a\}$, or (3) $\epsilon\cup\{a\}$ is incoherent and $\epsilon$ must be replaced by a `closest approximation' $\epsilon'$ containing $a$.\\

The entire process begins with switching $a$ to an excited state. The excitation propagates along $P$. Recall that for a coherent $\epsilon$, $\epsilon\cup\{a\}$ is incoherent iff there exists $b\in\epsilon$ such that $a\leq b^\ast$. Thus, under the operative assumption that $\epsilon$ is coherent (which may well be false), $\epsilon\cup\{a\}$ is proved to be coherent iff the contradiction detector raises no flags as a result of our exciting $a$. In this case $\epsilon$ will be replaced by $\epsilon\cup\{a\}$ -- turn $a$ on if it was off -- and we are done. In the situation when a contradiction is detected at $b\in P$ ($b\geq a$), simply turn $b^\ast$ off and turn $b$ on for all such $b$ in order to obtain $\epsilon'$.\\

To gain a better understanding of the actual meaning of this updating process, we examine it for the special case when $\epsilon$ is a vertex of $\Gamma(P)$ (that is: $\OO$ is completely decided -- though not necessarily right -- about the current state of the observed system). In this case it is easy to see that $\epsilon'$ is the unique vertex satisfying $\epsilon'\in I(\epsilon,v)$ for all $v\in V(a)$. Thus, in this case $\epsilon'$ is, in a sense, the best possible approximation of $\epsilon$ by elements of $V(a)$. Intuitively, we think there is no better candidate for $\epsilon'$. 

Now, it is unrealistic to assume that the propagation process takes no time, while it is reasonable to assume that the cells of the realization of $P$ (the nodes of the graph) have equal physical characteristics. Thus, the excitation signal will propagate through $P$ at a linear pace, implying our updating algorithm produces $\epsilon'$ in a time linear in the length of the longest maximal chain (in $P$) joining $a$ with $b$, where $b>a$ and $b^\ast\in\epsilon$. Clearly, this is as efficient as one may hope for. Of course, the propagation property plays the role of an extremely powerful parallel processor with the capacity to handle a potentially unbounded number of parallel computation threads.

\subsection{More realistic searching.}\label{subsection:more realistic searching} The reader will have noticed by now the similarities between our idea of propagation of excitation and the propagation of signals in a network of neurons. Consequently, the reader will have thought of the excessive optimism embodied in the assumption that the propagated signal does not dissipate: a neuron $\nu$ fires upon accumulating sufficient charge on its dendrites, and this charge is then distributed along the axons into the synapses; the more neurons have their dendrites connected to the axon of $\nu$, the less charge will accumulate on each of them. It is therefore reasonable to expect the signal to dissipate exponentially fast, if $P$ is sufficiently branched.\\

What does this mean for our updating algorithm?\\

First of all, it becomes possible for some of the elements $b\in\epsilon$ which satisfy $a<b^\ast$ to remain in $\epsilon'$ when the propagating excitation wave emanating from $a$ does not reach the corresponding nodes, thus failing to trigger the contradiction detectors.

As a result, $\epsilon'$ will not be coherent. This is the reason why we did not require coherence from $\epsilon$ in the definition of an observer, as well as our reason for emphasizing the possibility of $\epsilon$ being incoherent throughout the preceding discussion. Perhaps this is also the reason why we rarely observe humans with a completely coherent view of the world.\\

One should not completely despair of the task of ultimately updating $\epsilon$ with all the relevant elements of $P$, though. For let $a<b_1<\ldots<b_n=b$ be a maximal chain in $P$, where $b^\ast\in\epsilon$. Assuming that the observation $a$ remains valid for a length of time, there is a good chance that our observer will also make some of the observations $b_1,\ldots,b_m$, thus updating $\epsilon'$ with some of the $b_i$. This will increase the chance of the excitation wave propagating all the way to $b$ over {\it several} attempts at synchronizing memory with a recurring observation of of $a$. Learning requires persistence.\\

We want to remark that all the above does not diminish the efficiency of the algorithm: though instead of an accurate result the de-idealized algorithm only produces an approximation, the improvement in the quality of the recorded data is a linear function of the running time. Finally, we would like to turn the attention of the reader to the fact that the described updating process is, in fact, both an updating and a retrieval process. It seems natural to identify these two aspects of database management for a living organism.

\subsection{Other aspects of the proposed realization.}
We believe that the above analysis provides sufficient grounds for asking the question whether it is possible to realize a database structure as above using a neuronal network. If the answer is affirmative, then one should consider the way in which the other components of an observer $\OO$ could be realized by such a network. Here are some speculations in this direction:
\begin{description}
	\item[Realizing the excitation function $p$] Here is a na\"ive and partial approach to encoding the excitation function $p$, motivated by the idea of propagation of excitations through $p$: a neuron for $a$ can be viewed as supplying charge (through its axons) to every immediate successor $b\in P$. In order to force $b$ to fire as a consequence of $a$ firing, one needs to balance two parameters of this subsystem: (1) the action potential of the neuron $b$ and (2) the amount of charge delivered to the dendrites of $b$ through its connection with $a$. Tweaking this pair of parameters for every such pair $(a,b)$ will affect the range over which any excitation wave can propagate. Such a realization would also provide a tool for interpolating between `rigid binary thinking' and `probabilistic thinking', and motivates replacing the boolean algebra $\hsm{B}$ (appearing as the range of the realization map $f$ in the definition of an observer) by some other algebra, realizing other kinds of logic.
	\item[Structural updating] Neurons in brains of living organisms were observed in the process of creating/destroying synaptic connections while the observed animal was solving a problem. The ideas presented above poises the question about how restructuring of $P$ can be achieved through tweaking the excitation parameters of the system as we had just described. The next section discusses the logical aspects of the problem formally, but the question remains whether neuronal networks can be used to realize the proposed database management model, and even more importantly -- which cognitive phenomena can be simulated using such structures and how deep can one proceed with this analogy?
\end{description}

\section{Structural Updating of Observers.}\label{section:categorification}
\subsection{A deformation space for observers.}
There seem to be four components to update in an observer $\OO=(P,f,p,\epsilon)$. We have already discussed possible ways for altering $\epsilon$ in the context of the basic search problem, but that discussion required the structural component of the database -- the poc-set $P$ -- to remain constant. We must now address the problem of updating $P$.

Altering $P$ may involve the formulation of new questions, or a new realization that certain questions should be made comparable (or even equal) after being considered as transverse over some initial time period. Such alterations to $P$ will result in a re-definition of all the components of $\OO$. In just a few words, our idea is that the trigger for restructuring $P$ should be a significant change in the excitation function $p$: over time, some of the perceived events become negligible, and a clean-up operation is called for, to get the observer rid of the trouble of maintaining any record of such events. Changes in the values of the excitation function are external to the algorithmic structure of memory: these we assume to be generated by the physical organism, as a direct reaction to observations made by the sensors it possesses.\\

Structural alterations should occur as a result of sensors being added, increased precision in existing sensors, communication with other observers and other observations causing a re-evaluation of excitation levels. In other words, alterations of the poc-set structure underlying the memory graph should correspond to a change in the natural language used by the observer. More precisely, we note that distinct observers may share common questions and be well aware of this fact. A formal way of expressing this in a general context is to have all questions `tagged' by symbols from a prescribed alphabet $\fat{A}$, and all the participating agents using the same set of tags.\\

It is convenient for $\fat{A}$ to be an infinite set. For every subset $A$ of $\fat{A}$, let $D(A)$ denote the set of symbols $A\sqcup A^\ast\sqcup\{0,0^\ast\}$ endowed with the complementation operator given by $a\mapsto a^\ast$ for all $a\in A\sqcup\{0\}$ and $a^\ast\mapsto a$ for all $a\in A^\ast\cup\{0^\ast\}$. Let $\Poc{A}$ denote the set of all poc-set structures of the form $(D(A),\leq,\ast)$ for which the symbol $0$ is the minimum element. By $\PocA$ we mean the union of $\Poc{A}$ over all finite subsets $A$ of $\fat{A}$. For $P\in\Poc{A}$, we shall say that $D(A)$ is the \emph{support} $\supp{P}$ of $P$.\\


Denote by $\square^P$ the set of all functions $q:V\Gamma(P)\to[0,1]$ satisfying $\sum_{u\in V\Gamma(P)}q(u)=1$. $\square^P$ is a standard Euclidean $\left(\card{V\Gamma(P)}-1\right)$-dimensional simplex. To any pair $(P,p)$ with $P\in\PocA$ and $p:V\Gamma(P)\to[0,\infty)$ non-zero, we associate the point 
\begin{displaymath}
	\PR{\OO}{p}=\frac{p}{\displaystyle \sum_{u\in V\Gamma(P)}p(u)}\in\square^P\,.
\end{displaymath} 

\begin{figure}[t]
	\centering{\includegraphics[width=\textwidth]{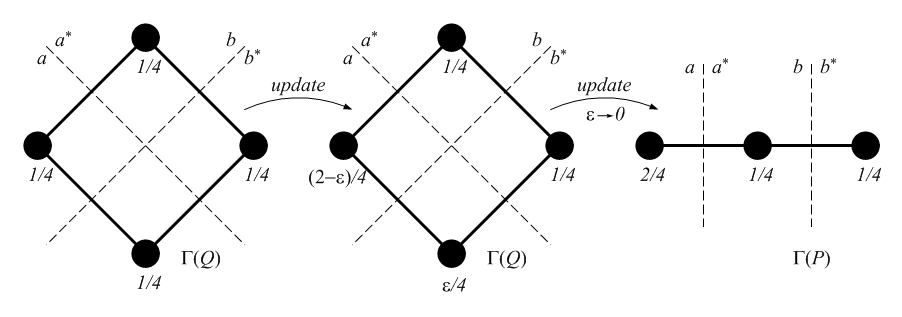}}
	\caption{\protect\scriptsize A simple example of degeneration of a memory graph due to recalculation of probabilities.\protect\normalsize
	\label{fig:degeneration}}
\end{figure}

Let us consider what happens as $q\in\square^P$ is being moved toward the boundary of $\square^P$: in the eye of an observer $\OO$ with underlying poc-set $P$ and $\PR{\OO}{p}=q$, the events corresponding to diminishing values of $q$ gradually become negligible; in the limit (as $p$ reaches a face $F$ of $\square^P$), $\OO$ will ignore such events, treating them as irrelevant. As a result, some questions may become redundant in the eyes of $\OO$ or the relations between them might change. However, we note that approaching the boundary of $\square^P$ from inside $\square^P$ will never result in a pair of nested elements of $P$ becoming transverse, while a pair of transverse elements may transform into a nested pair or even into a pair of equal/complementary elements. The resulting `degenerate' poc-set should then correspond to the face $F$. A convenient notion in this context is:
\begin{defn}[Corners] Suppose $P$ is a finite poc-set. A {\it corner} of $\Gamma(P)$ is the full subgraph of $\Gamma(P)$ induced by a set of vertices of the form $V(a,b)$ for some proper elements $a,b\in P$.
\end{defn}
Putting the preceding discussion in other words, a degeneration of $P$ should occur whenever $\PR{\OO}{p}$ assigns a negligible value to some corner of $\Gamma(P)$. Here is a very simple example to keep in mind:
\begin{example}\label{example:degeneration} Recall example \ref{example:injective but not an embedding}, where we had two poc-sets $P$ and $Q$ each generated by a pair of distinct proper elements $a,b$ so that $a\pitchfork b$ in $Q$ and $a<b$ in $P$. Then $\Gamma(Q)$ is a $4$-cycle, while $\Gamma(P)$ is an interval with three vertices -- see figure \ref{fig:degeneration}, left-hand side.

Assigning equal excitation probabilities to the four vertices of $\Gamma(Q)$ places the corresponding weighted graph at the barycenter of $\square^Q$. Now consider a perturbation of those probabilities as described in the central diagram of figure \ref{fig:degeneration}: the excitation probability of the corner $V(a,b^\ast)$ vanishes as $\epsilon$ approaches zero, leading to a restructuring of the poc-set $Q$ into the poc-set $P$.
\end{example}
\begin{defn}[Degeneration] If $P,Q\in\PocA$, say that $Q$ degenerates into $P$ (denoted by $P\leq Q$) iff $\supp{P}\subseteq\supp{Q}$ and there exists a {\it retraction} of $Q$ onto $P$: a poc morphism $r:Q\to P$ restricting to the identity on $\supp{P}$.
\end{defn}

Using this notion, we are now able to glue all the simplices $\square^P$, $P\in\PocA$ to obtain a parameter space encoding the relationships among all the graphs $\Gamma(P)$ supported on $\fat{A}$:
\begin{defn}[The Deformation Space of Observers over $\fat{A}$]\label{deformation space} For every $P,Q\in\PocA$ satisfying $P\leq Q$ and for each retraction $r:Q\to P$, we define an injective map $\delta=\delta_r:\square^P\to\square^Q$ by setting
\begin{equation}
	\left(\delta(p)\right)(u)=\left\{\begin{array}{rl}
		p(v)	&\mathrm{if}\; u=r^\circ(v)\\
		0	&\mathrm{if}\; u\notin Im\left(r^\circ\right)
	\end{array}\right.
\end{equation}
for every $u\in V\Gamma(Q)$. We then construct a space $\hsm{U}=\hsm{U}(\fat{A})$ --- {\it the deformation space of observers over $\fat{A}$} --- as the quotient of the union $\bigcup_{P\in\PocA}\square^P$ by the equivalence relation generated by the relations of the form $p\equiv \delta_r(p)$ for all $p\in\square^P$, all $P,Q\in\PocA$ and all retractions $r:Q\to P$.

For every $p\in\hsm{U}$, denote by $\ch{p}$ the unique simplex of $\hsm{U}$ containing $p$ in its interior. Equivalently, $\ch{p}$ is the top-dimensional simplex of $\hsm{U}$ containing $p$.
\end{defn}
Thus $\hsm{U}$ is obtained from the collection of all simplices of the form $\square^P$ by identifying some of them along faces (gluing). The main characteristic of the space $\hsm{U}$ is that any path in this space corresponds to a process of altering some memory graph through introduction of questions (moving away from a face of simplex) or removal/identifications of/among redundant questions. Also note that any two simplices $\square^P$ and $\square^Q$ occur as faces of $\square^R$, where $R$ is the poc-set with support $\supp{P}\cup\supp{Q}$ and no relations (every pair of proper questions is transverse). This shows that $\hsm{U}$ is the result of carrying out face identifications on one infinite-dimensional simplex.

\subsection{The Updating Postulate.}\label{subsection:updating postulate} Up till now we have not made any assumption regarding the way in which the updating of an observer occurs, leaving the model static: at any point in time, an abstract observer can be associated with a real-life observer by, say, freezing the latter and studying the poc-set structure of the contents of its memory. Of course, it is the evolution of an observer that is of interest in relation to studying learning processes.\\

Consider the postulate: {\it If a sequence $\OO_n=(P_n,f_n,p_n,\epsilon_n)$ ($n\geq 0$) represents consecutive stages in the evolution of the same observer, then, for every $n\geq 1$, either $\ch{p_{n-1}}$ is a face of $\ch{p_n}$ or $\ch{p_n}$ is a face of $\ch{p_{n-1}}$}.\\

This postulate means, essentially, that, whatever the implementation of any given observer, and whatever procedure is used for updating it, any such updating results either in a degeneration as described above, or in an `inverse degeneration'. Thus, one could imagine the evolution of a memory graph to be an alternating sequence of expansion/contraction moves.

The reason for stating such a postulate is that of economy: an attempt should be made to free the observer of the burden of maintaining low-priority vertices in $\Gamma(P_{n-1})$ whenever the same information can be encoded by degenerating $P_{n-1}$ into a poc-set $P_n$ with a smaller dual graph. Since $\Gamma(P)$ tends to grow exponentially with the size of $P$, this optimization becomes necessary due to its potential for conservation of resources.\\

The point of view described above provides a possible explanation of why humans, for example, find it so hard to update a set point of view (on practically any given issue), or, more generally, why good studying of a recurring phenomenon results in automatisms that often bar the student from creatively responding to an unpredicted change of circumstances. Looking again at the example in figure \ref{fig:degeneration}, it is easy to imagine the amount of destruction resulting from a more complex degeneration process (occurring in a more complex graph). It is then stunning to see how simple it is to achieve this kind of update (e.g., given an implementation such as the one proposed in section \ref{section:implementation}, this kind of update can be achieved through rewriting just a few pointers and then collecting the garbage), compared to how hard it is (algorithmically) to build a new hierarchy of related questions, perhaps reconstructing parts of the old one, and then shaping the graph (through continued observation leading to degenerations) so that it fits a more complete picture of the observed reality.

While the price to pay for mistakingly erasing low-priority nodes is great, the benefit of eliminating objective redundancies may outweigh this price if the threshold for erasing a low priority vertex is low enough.\\

To summarize, we have just described how the updating postulate turns the processes of updating memory, logical deduction and forgetting into different aspects (depending on context) of the same principle, by restricting the effect of an `elementary updating process' on the memory structure.

\subsection{Structure of $\hsm{U}$ {\it vs.} Natural Language.}\label{subsection:natural language} Let us discuss an additional aspect of the idea of degeneration in the memory graph of an abstract observer. Suppose now that $\epsilon>0$ is given with the property that every observer $\OO=(P,\ldots)$ in a given group of observers considers a vertex $u\in V\Gamma(P)$ negligible if $\PR{\OO}{u}<\epsilon$. Once again, let $\fat{A}$ be the alphabet of questions recognized by the members of the group.

For any $n\geq 0$, let $\hsm{U}^{(n)}$ denote the $n$-skeleton of $\hsm{U}=\hsm{U}(\fat{A})$ -- the union of all simplices of $\hsm{U}$ having dimension $n$ or less.\\

Then, whenever $\left|V\Gamma(P)\right|$ exceeds $\frac{1}{\epsilon}$ (for some $P\in\PocA$), a whole ball about the baricenter of the simplex $\square^P$ becomes irrelevant for the discussion of this particular group of observers. This makes the $\frac{1}{\epsilon}$-skeleton of $\hsm{U}$ much more relevant to the discussion of `language' in the given group than the space $\hsm{U}$ itself. Moreover, replacing $\hsm{U}$ by $\hsm{U}^{(1/\epsilon)}$ in the role of a `space of all observers' increases its topological complexity, which may also be relevant in the discussion of language structure and formation.\\

Here is a more precise formulation of what we mean by this. Consider the process of parents teaching their newborn child. The newborn has many `questions' available, but not much can be made out of them at the very beginning: they are not ordered yet. The environment provides a lot of input, and the role of the parents is to serve as a filter, protecting the child from input it is yet incapable of processing. The result is that the parent slowly synchronizes the child's memory structure with their own, increasing the chances that the child's sensors carry the same meaning as analogous sensors in the parent. As a result, the individuals in a stably evolved population will have many common patterns in their memory structures: a vast majority of individuals will agree on certain associations between different inputs (consider our attaching a very specific meaning to the sound of the word `green', unless we are color-blind). Thus, for any given population, a subset $\fat{B}$ of $\fat{A}$ exists, for which a poc-set structure is already decided among the adults of that population, and all discourse and synchronization among adults is confined to the sub-complex of $\hsm{U}(\fat{A})$ -- denote it by $\hsm{U}(\fat{A},\fat{B})$ -- constructed using only those $P\in\PocA$ for which the poc-set structure on $\supp{P}\cap\fat{B}$ inherited from $P$ coincides with the poc-set structure inherited from $\fat{B}$. The evolution of the population's language is then partially described by the evolution of $\hsm{U}(\fat{A},\fat{B})$ over time; the learning goal of a child in the population is to synchronize their memory structure with that defined by the parent population.\\

In subsection \ref{subsection:common sense} we discussed the effect of the capability for employing logical (and perhaps other) connectives for enriching the poc-set structure of an observer by creating `new questions from old'. The main observation was that an observer needs to balance between two extremes: over-using this capability leads to exponential inflation of the memory structure making it impossible to maintain efficiently; not using it enough prevents the memory structure from attaining a sufficient level of refinement and blocks the observer from capturing more subtle information about the observed system.

It is easy to see how this idea gets incorporated into the discussion of the deformation space $\hsm{U}(\fat{A})$: simply impose additional algebraic structure on $\{0,0^\ast\}\cup\fat{A}\cup\fat{A}^\ast$. For example, here is a natural way of adding Boolean connectives to the alphabet. For each $a,b\in\fat{A}$ one assumes that the symbols of the form $a\wedge b$, $a\vee b$ (and all incident finite formulae) are in the alphabet, together with all the ensuing relations, and we require all admissible poc-set structures to be synchronized with these symbols -- e.g., for all $a,b,c\in P\in\PocA$ require that $a\wedge b<a$, $a\wedge b<b$ and $(c<a)\wedge(c<b)$ must imply $c<a\wedge b$. Yet again, the result is a sub-complex of the original $\hsm{U}(\fat{A})$. The two refined constructions we had just discussed can be combined together or compared -- any option will yield an aspect of the phenomenon one can only call by the name {\it the language spoken by the population}.

\section{Discussion}\label{section:discussion}
We have defined and, to an extent, analyzed a family of databases designed to maintain the memory of an arbitrary entity observing the evolution of an environment by means of binary sensors. Motivated by the example of living organisms, we assumed the following:
\begin{itemize}
	\item[-] The observer evolves with time, possibly `growing' new sensors;
	\item[-] The observer may interact with the environment;
	\item[-] The observer has mechanisms to assign excitation levels to observed data -- we refer to these collectively as the observer's {\it evaluation mechanism}.
\end{itemize}
Our main goal was to construct a memory system capable of dealing with arbitrary input, given the guidance provided by the evaluation mechanism. This means input is not treated according to its objective significance to the observer, but rather according to the {\it subjective} importance assigned to the input by the evaluation mechanism. When the evaluation mechanism and reality are in tune with each other, input from the environment contributes to a more accurate record of events in the database, and to better decision-making on the part of the observer.

\subsection{The Invariance Principle and alternate logics.} A major trait of our model is a clear separation of the algorithmic side (the database) from the physical side (the realization). On the former side, an observer $\OO=(P,f,p,\epsilon)$ is required to maintain the poc-set structure $P$, the (conjectural) current state $\epsilon$ and the graph $\Gamma(P)$ with vertices weighted by $p$. On the latter side, we have the poc-morphism $f:P\to\hsm{B}$ (recall $\hsm{B}$ is the Borel $\sigma$-algebra on $X$) {\it objectively} representing the observer's sensors. Together, these components produce a picture of the mind of a real-life observer, frozen in a given moment of its evolution. The evaluation mechanism and actual physical realization of this system are given the task of effecting the transition of $\OO$ from any given current state to its next state. This is done through:
\begin{enumerate}
	\item\label{num:sampling} Sampling sensory input. In this context, the physical realization of the observer is considered a part of the observed environment.
	\item\label{num:current state} Updating the current state record given by $\epsilon$ to fit the most recent observations (see section \ref{section:implementation})
	\item\label{num:evaluation} Re-evaluating excitation levels relevant to the current observations.
	\item\label{num:restructuring} Optimizing the database structure $\Gamma(P)$ to fit the new excitation levels.
\end{enumerate}
Note how clear cut is the division of labor between the algorithmic and physical components of the model: stages (\ref{num:sampling}) and (\ref{num:evaluation}) are totally dependent of the physical organism, while stages (\ref{num:current state}) and (\ref{num:restructuring}) may be formulated completely in terms of the algorithmic structure, only depending on the physical `shell' for their input and execution of the algorithmic tasks at hand.

In section \ref{subsection:general considerations} we had already mentioned the more obvious potential of such a `two-lobe' approach. In a manner of speaking, this approach achieves a separation of `software' from `hardware', enabling the study of the question: what kind of hardware is capable of supporting this software?\\

Note that the representation map $f$ is the only objective component of an observer $\OO$. The choice of range for $f$ (the Boolean algebra $\hsm{B}$) corresponds to the assumption that $\OO$ uses binary sensors to observe its environment. As a result, a deeper aspect of the invariance principle comes to light: one could weaken the binarity assumption considerably by replacing the range of $f$ by a larger algebra quantifying over $X$. For example, one could use the space $\Psi(X,\mu)$ of probability distributions on $(X,\mu)$ as a range for $f$. Indeed, the space $\Psi(X,\mu)$ has a fuzzy logic underlying it, while it still retains a poc-set structure induced from $\hsm{B}$: for every $\psi,\psi'\in\Psi(X,\mu)$ one defines $\psi^\ast=1-\psi$, and $\psi\leq\psi'$ iff $\psi(x)\leq\psi'(x)$ holds throughout $X$. This change opens new horizons we have not explored yet, as it greatly alters the interpretation of the database. To be precise: the observer $\hsm{O}$ {\it observes} fuzzy phenomena, but {\it records} them as if they were deterministic. 

Quantum logics and temporal logics are of particular interest to us in this context: replacing $\hsm{B}$ by a quantum logic may be expected, according to \cite{[Kauffman-LOF]}, to enable $\OO$ to consider self-referential statements; a logic with a temporal structure could allow $\OO$ to become naturally aware of the passage of time, and consequently -- aware of its own learning processes.

At the same time, while the range of the representation map is allowed to vary, it is not necessary to change the structure of the database itself. It is then the {\it interpretation} of the database structure and its role in the interaction between the database and the physical realization that changes when the binary logic of our model is replaced by another logic. 
This opens the door to questions regarding the nature of our own (e.g., human) `guessing algorithms', and motivates one to ask whether a memory structure such as the one we are proposing in this paper can be used to enforce decision making by an autonomous agent in non-deterministic situations.

\subsection{Explanatory power of the model} The introduction strongly emphasizes the idea that frequently observed weaknesses of the human thought process, including the destructive process of forgetting, should serve as testing stones for any model contending for the high title of ``model of the memory of living organisms''. More precisely: any contender must be capable of demonstrating as many such phenomena as possible to be different aspects of its {\it normal} operation.

Now that all the proper language has been developed and all modeling assumptions have finally been stated, we can summarize the most notable achievements of our model on this front. Put in our new language, the memory structure of an organism in an $\fat{A}$-speaking population is modeled by a sequence $(\OO_n)_{n\in\ZZ}$ of observers $\OO_n=(P_n,f_n,p_n,\epsilon_n)$. 
Recall that structural updating is carried out in accordance with the updating postulate from \ref{subsection:updating postulate}, so for every $n\geq 1$ we have either that $\ch{p_{n-1}}$ is a face of $\ch{p_n}$ or that $\ch{p_n}$ is a face of $\ch{p_{n-1}}$ in the deformation space $\hsm{U}(\fat{A})$.
\subsubsection*{Permanent loss of information.} There are two ways in which information is lost by our database structure. Permanent loss of information occurs whenever the transition $\OO_n\to\OO_{n+1}$ involves a degeneration of a corner $V(a,b)$ of $\Gamma(P_n)$ whose representation in $\hsm{B}$ has non-zero probability: our observer will, never the less, consider the logical implication $a<b^\ast$ as a property of $X$ unless renewed observations somehow force a reverse structural update. There are two important observations to make in this context:
	\begin{enumerate}
		\item `Forgetting' the event in $\hsm{B}$ corresponding to the corner $V(a,b)$ is equivalent to committing the statement $a<b^\ast$ to memory. Therefore, every act of memorization -- every act of recording a conclusion about the structure of $X$, if you like -- may, potentially, result in loss of information. However, structural updating is necessary for keeping the database from blowing up, so this price has to be paid whether we like it or not.
		\item Information lost (by the database) in this way is very hard to recover if we realize the database in a manner such as the one described in section \ref{section:implementation}. The reason is that -- at least in the ideal implementation -- recording the relation $a<b^\ast$ implies that whenever the observation $a$ is made, propagation of excitation implies the observation $b^\ast$ is made as well. In the non-ideal implementation, it is possible that the observation $b^\ast$ will not be made if there are sufficiently many $x\in P$ satisfying $a<x<b^\ast$: dissipation of the excitation signal may cause the search algorithm not to reach $b^\ast$ at all. In either case, an incoming observation of $b$ {\it and} $a$ may be simply discarded by the system unless the evaluation mechanism gives this input an excitation value that is too high to ignore. Possibly, this issue may be better understood in the context of a more realistic implementation of the database by, say, a neuronal network.
	\end{enumerate}  
\subsubsection*{Temporary loss of information.} We consider situations when the information sought for is contained in the database, but, for some reason, is hard to find. Formally speaking, the system is being fed an observation $a\in P$ (or observations $a_1,\ldots,a_n\in P$), and is expected to produce the conclusion $x\in P$ -- to respond to the observation in a way that provides proof (to the environment, in some sense) that it has `understood' the implication `$a$ implies $x$'; however, the desired demonstration fails to occur. 
	
In the context of an implementation based on propagation of excitation, these situations are easy to explain/recreate: an input, or combination of inputs, triggers a cascade of `internal' observations, due to propagation of the excitation signal(s); if the signal dissipates before the node $x$ is reached, then the observer will fail to produce the required reaction. However, repeated queries along a path (in $P$) from $a$ to $x$ may result in the excitation wave reaching $x$ {\it eventually}.
\subsubsection*{Bound on parallel processing.} This is a point we have hardly touched in this paper, but would like to point it out as a motivation for future research. If the network of nodes realizing $P$ can be excited at several nodes $a_1,\ldots,a_n$ simultaneously, corresponding to the observation $a_1\wedge\ldots\wedge a_n$ being made, then the range of reachable conclusions potentially increases with $n$. The observation $a_1\wedge\ldots\wedge a_n$ serves as proof that $\{a_1,\ldots,a_n\}$ is a coherent family,  which means $V(a_1,\ldots,a_n)\subset\Gamma(P)$ is non-empty. If there is some {\it a priori} upper bound on $n$ (an upper bound on the number of simultaneous observations), then, as the set of available sensors grows, the ability of the observer to relate to specific vertices of $\Gamma(P)$ diminishes unless the organism has a tool for introducing sensors corresponding to conjunctions of families of other sensors. In any case, a bound on the `bandwidth' of the implementation of the poc-set $P$ automatically implies a restriction on the ability of the organism to deal with complex input.
	
	Some organisms, like modern humans, have developed tools to help them circumvent such difficulties. We are able to articulate the results of our thought process and write them down, thus creating an artificial feedback loop with its own memory capacity. Reading our text at a later time re-excites thought processes that were abandoned earlier, which, armed with new data, updated memory and new notions may now have a better chance of reaching the desirable goal.
	
\subsubsection*{Noise.} Remember that song that you cannot stop humming? The gossip information from television that you recall with an ease igniting the envy of all the material you have read for work and successfully forgotten? How about the dead silence, or the chirping birds, or deafening heavy metal music that you need for concentration? 
	
	Our model explains these through `bandwidth' considerations from the preceding paragraph and through known properties of the evaluation mechanism in living organisms. For us and for many other animals, the most significant component of our excitation levels seems to be physiological stress.
	
	Physiological stress plays the same role in biology as temperature in physics: when stress is totally absent, the animal is overly docile and disinterested with its environment; when there is too much stress, metabolic reactions peak and the animal becomes incapable of reacting to the environment due to the total chaos in the input it has to process. Depending on the processing mechanism, some inputs (normally weak, periodic, predictable) will stress an animal just enough to cause it to ignore most common but irregular minuscule sources of stress (distractions), allowing it to use the rest of its capacity for parallel processing for quality `thinking' (this is a state of `calm' for the animal). Other inputs get evaluated so high, that they keep occupying the animal's computational resources for long periods of time -- at least as long as the inputs persist -- which results in wasting computational resources on the evaluation and re-evaluation of those signals and any `memories' they may trigger as a result of the searching process (as described in section \ref{section:implementation}). Reliving these memories, in turn, may result in more stress, creating a cycle that only breaks with the introduction of a much more powerful signal or with the disappearance of the majority of the stressing factors from the observer's horizon.
	
	This is, of course, a speculative picture, but we find the simplicity of the explanation appealing.	In particular, our model then serves as a (formal, mathematical) link between the algorithmic and psychological aspects of learning.

\subsection{Biological connections and natural language.}
The preceding review started with computational aspects of information processing (as realized in the proposed model) and ended with the biological ones. In the context of living beings (which are forced to interact with their envirnoment), neither can be discussed separately from the notion of a natural language. Different organisms have evolved their information processing tools to different levels (different parts of the human nervous system may be traced back to different stages of the planetary evolutionary process), and the same can be said about the evolution of their means for communicating with the environment. Thus, inevitably, the evolution of natural languages is related to the evolution of physical realizations of memory (e.g., brains) and of the algorithmic tools maintaining the underlying database structures.\\

The current textbook definition of a natural language is somewhat lacking in mathematical rigor. Definitions such as ``by a natural language we mean human languages such as English, Spanish, Arabic etc.'' are generally accepted, but can hardly be considered rigorous.\\

We believe our approach has a new bearing on the formal understanding of the notion of natural language, as well as the process of language acquisition at the level of the individual learner. Niyogi (see \cite{[Niyogi]} section 1.1, p.16) provides a learnability argument in favor of restricting the family of languages human learners are capable of producing: 
\begin{itemize}
	\item[] ``The necessary and sufficient conditions for successive generalization by a learning algorithm has been the topic of intense investigation by the theoretical communities in computer science, mathematics, statistics and philosophy. They point to the inherent difficulty of inferring an unknown target from finite resources, and in all such investigations, one concludes that {\it tabula rasa} learning is not possible. Thus children do not entertain every possible hypothesis that is consistent with the data they receive but only a limited class of hypotheses. This class of grammatical hypotheses is the class of possible grammars children can conceive and therefore constraints the range of possible languages that humans can invent and speak.''
\end{itemize}
The mathematical nature of these constraints remains unclear unless a reasonable formal universal model of a learner is provided. In some simplified sense, this is precisely what our model is. Moreover, we show that it comes equipped with a tool for formally defining the learning goal for language acquisition. In what follows, we will try to substantiate this claim using the `deformation spaces of observers' we have recently considered in section \ref{section:categorification}.\\

In the context of humans, the notion of a natural language is directly associated with the use of sound for the articulation of one's thought process. However, once the use of signs and signals is permitted -- written language, for example -- one has to face a growing number of alternatives stretching the definition far beyond the realm of the audible. For example, where should one place sign language? by what means should one communicate with the little green deaf visitors from Mars? If one is to take into account so many variations on the original notion of natural language, then surely this notion should apply to every self-sufficient coherent system of signals allowing the exchange of information among sentient beings. 

Stop. At this point, it will do us good to realize our inability to discern sentient beings from non-sentient ones. At best, we are able to distinguish solitary species from social ones. But perhaps this is where one finds the key to solving our problem: while {\it any} repetitive pattern of information exchange among physical entities may be considered a language (in a broader understanding of the term), we should be seeking a notion that is inherently linked with shared territory and ordered communication among organisms. It is no big surprise that humans and human languages have evolved from social primates rather than from solitary wasps, say. After all, from the evolutionary point of view, populations of solitary species have no use for complex articulated information exchange, while populations of social organisms may benefit from this ability. 

It is this last idea that motivated the construction of a deformation space $\hsm{U}(\fat{A})$ of $\fat{A}$-speaking observers. The space $\hsm{U}(\fat{A})$ is best understood in the context
of a population $\hsm{P}$ of organisms sharing a set $\fat{A}$ of statements about $X$. The elements of $\fat{A}$ should not be thought of as corresponding to words in the language spoken by the population, but rather to the {\it shared meanings of complete sentences}. An outside observer trying to study $\hsm{P}$ may initially be oblivious of the different meanings of the elements of $\fat{A}$, but they could try to guess those meanings from contextual data collected while observing the evolution of memory structures of members of $\hsm{P}$. Similarly, a new member of $\hsm{P}$ -- a newborn child, say -- has to observe other members of $\hsm{P}$ and communicate with them in order to uncover the actual meaning of each element of $\fat{A}$.

Now, as we have already pointed out in section \ref{subsection:natural language}, shared meaning implies shared logical structure, so it is reasonable to assume that the adult part of the population will have a shared record of certain relations among elements of $D(A)=\{0,0^\ast\}\sqcup\fat{A}\sqcup\fat{A}^\ast$, which, inevitably will show up in their memory structures. The young of $\hsm{P}$ can then be considered as trying to restructure their memory graphs accordingly. It is also plausible to expect various connectives to appear as tools for enriching the set of statements meaningful to the population: this introduces an algebraic structure on $D(A)$, e.g. conjunction and implication operators in the case of human languages. More generally, {\it any} shared structuring of the information exchange among members of $\hsm{P}$ ends up as a shared property of the adult memory structures that has to be learned (through restructuring) by the younger members. We have shown in \ref{subsection:natural language} how additional structural properties of $D(\fat{A})$ (or subsets of the form $D(\fat{B})$, $\fat{B}\subset\fat{A}$) define special subspaces $\hsm{U}(\fat{A},\fat{B})$ of the the deformation space $\hsm{U}(\fat{A})$ with non-trivial topology derived from the properties of $\fat{B}$. Mathematically speaking, studying the topology and geometry of these subspaces corresponds to studying the language `spoken' by $\hsm{P}$, from an outsider's viewpoint. From an insider's viewpoint, a learner's goal is to have its memory structure synchronized with that of the entire population: the population provides input consistent with $\fat{B}$ while the learner's evaluation mechanism provides guidance directing the evolution of the learner's memory structure towards $\hsm{U}(\fat{A},\fat{B})$; inadequate guidance (e.g., the learner was born autistic) may prove insufficient for the learner to achieve the desired result; inadequate pressure by the parent population (input contradicting the learner's evaluation mechanism) may result in failure to acquire the language of the population.\\

At the current stage, this last observation is not of much use to a linguist studying a specific language. However, the novelty in our point of view is in that we have shown how language acquisition occurs (in the framework of our model) as a result of a learning process that is {\it not} guided by a specific pre-defined problem, but by the very general problem of the learner (e.g., a child) attempting coherent communication with its immediate environment. The only guidance present is the {\it subjective} guidance by the evaluation mechanism, and this can be swayed in many different directions by an attentive teacher. 

To summarize, all the above suggests that the constraints Niyogi discussed may be derived from two sources. The first is an accurate description of the evaluation mechanism providing the guidance for the learning process. The second is a good description of the logic employed by the learner and the way in which it is realized by the updating algorithms of the learner's memory structure. While our model assumed classical Boolean logic, our discussion of the invariance principle and its realization in our model shows how this assumption can be lifted. We leave this direction to future research.

\providecommand{\bysame}{\leavevmode\hbox to3em{\hrulefill}\thinspace}
\providecommand{\MR}{\relax\ifhmode\unskip\space\fi MR }
\providecommand{\MRhref}[2]{%
  \href{http://www.ams.org/mathscinet-getitem?mr=#1}{#2}
}
\providecommand{\href}[2]{#2}

\end{document}